\documentclass{article}

\usepackage{microtype}
\usepackage{graphicx}
\usepackage{subfigure}
\usepackage{booktabs} 

\usepackage{hyperref}

\usepackage[accepted]{icml2023}

\usepackage{amsmath}
\usepackage{amssymb}
\usepackage{mathtools}
\usepackage{amsthm}

\usepackage[capitalize,noabbrev]{cleveref}

\theoremstyle{plain}
\newtheorem{theorem}{Theorem}[section]
\newtheorem{proposition}[theorem]{Proposition}
\newtheorem{lemma}[theorem]{Lemma}

\theoremstyle{definition}
\newtheorem{definition}[theorem]{Definition}

\theoremstyle{remark}

\usepackage[textsize=tiny]{todonotes}

\usepackage{multirow}

\newcommand{\eg}{e.\,g., }
\newcommand{\ie}{i.\,e., }

\newcommand{\oms}{\{\!\!\{}
\newcommand{\cms}{\}\!\!\}}
\newcommand{\Bigoms}{\Big\{\!\!\Big\{}
\newcommand{\Bigcms}{\Big\}\!\!\Big\}}

\icmltitlerunning{Path Neural Networks: Expressive and Accurate Graph Neural Networks}

\begin{document}

\twocolumn[
\icmltitle{Path Neural Networks: Expressive and Accurate Graph Neural Networks}



\icmlsetsymbol{equal}{*}

\begin{icmlauthorlist}
\icmlauthor{Gaspard Michel}{equal,polytechnique,deezer}
\icmlauthor{Giannis Nikolentzos}{equal,polytechnique}
\icmlauthor{Johannes Lutzeyer}{equal,polytechnique}
\icmlauthor{Michalis Vazirgiannis}{polytechnique}
\end{icmlauthorlist}

\icmlaffiliation{polytechnique}{LIX, \'Ecole Polytechnique, IP Paris, France}
\icmlaffiliation{deezer}{Deezer Research, Paris, France}

\icmlcorrespondingauthor{Gaspard Michel}{gmichel@deezer.com}

\icmlkeywords{Machine Learning, ICML}

\vskip 0.3in
]



\printAffiliationsAndNotice{\icmlEqualContribution} 

\begin{abstract}
Graph neural networks (GNNs) have recently become the standard approach for learning with graph-structured data.
Prior work has shed light into their potential, but also their limitations.
Unfortunately, it was shown that standard GNNs are limited in their expressive power.
These models are no more powerful than the $1$-dimensional Weisfeiler-Leman ($1$-WL) algorithm in terms of distinguishing non-isomorphic graphs.
In this paper, we propose Path Neural Networks (PathNNs), a model that updates node representations by aggregating paths emanating from nodes.
We derive three different variants of the PathNN model that aggregate single shortest paths, all shortest paths and all simple paths of length up to $K$.
We prove that two of these variants are strictly more powerful than the $1$-WL algorithm, and we experimentally validate our theoretical results.
We find that PathNNs can distinguish pairs of non-isomorphic graphs that are indistinguishable by $1$-WL, while our most expressive PathNN variant can even distinguish between $3$-WL indistinguishable graphs.
The different PathNN variants are also evaluated on graph classification and graph regression datasets, where in most cases, they outperform the baseline methods.
\end{abstract}

\section{Introduction}
Graphs have emerged in recent years as a powerful tool for representing irregular data.
Among many other applications, graphs have been used to model the relationships between individuals within a social network~\cite{easley2010networks} and the interactions between the atoms of a molecule~\cite{stokes2020deep}.
The large availability of graph-structured data has motivated the development of machine learning algorithms that are designed for this kind of data.
Notably, most of those algorithms belong either to the family of graph kernels~\cite{kriege2020survey,borgwardt2020graph,nikolentzos2021graph} or to that of graph neural networks (GNNs)~\cite{wu2020comprehensive,zhou2020graph}.
Due to some limitations that are inherent to kernels (\eg they do not scale to large datasets, they struggle with continuous features, etc.), GNNs have become the most common approach for dealing with graph learning problems.

GNNs have been studied extensively in the past years.
So far, research in the field has mainly focused on message passing architectures~\cite{gilmer2017neural}.
These models follow a recursive neighborhood aggregation scheme where each node aggregates the representations of its neighbors along with its own representation to compute new updated representations.
It has been shown that there is a connection between the neighborhood aggregation scheme of these models and the relabeling procedure of the Weisfeiler-Leman (WL) algorithm, a well-known heuristic for testing graph isomorphism~\cite{xu2019powerful,morris2019weisfeiler}.
More importantly, it was shown that the standard message passing architectures are at most as powerful as the WL algorithm in terms of distinguishing non-isomorphic graphs.
This has led to the development of more complex models that focus on subgraphs~\cite{maron2019invariant,maron2019provably,cotta2021reconstruction}.
Some of these models are inspired by higher-order variants of the WL algorithm~\cite{morris2019weisfeiler,morris2020weisfeiler}.

Besides subgraphs, there are also other structures of graphs that can improve a model's expressive power.
For instance, paths can distinguish connected from disconnected graphs, while the WL algorithm fails to distinguish this property.
Unfortunately, finding all paths in a graph is NP-hard. 
However, some subsets of paths can be computed in polynomial time.
For instance, computing shortest paths in a graph is a problem solvable in polynomial time~\cite{dijkstra1959note}.
It has been shown that models that use shortest path distances between nodes as features can provide more expressive power than the WL algorithm~\cite{li2020distance}.
Furthermore, if we restrict the length of the paths to some small integer $k$, the parameterized complexity of computing all the bounded length paths is tractable.

In this paper, we propose Path Neural Networks (PathNNs), a GNN that generates node representations that are based on paths emanating from the nodes of graphs.
By computing different subsets of paths, we derive different variants of the proposed model.
We focus on subsets whose computation is possible in polynomial time, namely shortest paths and simple paths of bounded length.
For each path length, the proposed model uses a recurrent layer to encode paths into vectors and then, the representations of all paths emanating from a node are aggregated to produce the node's new representation.
We show that two of the three PathNN variants are strictly more powerful than the WL algorithm in terms of distinguishing non-isomorphic graphs.
Our theoretical results are confirmed by experiments on synthetic datasets that measure the models' expressive power.
Furthermore, we evaluate the PathNNs on real-world graph classification and regression datasets\footnote{Code available at \url{https://github.com/gasmichel/PathNNs_expressive}}.
Our results demonstrate that the different PathNN variants achieve high levels of performance and outperform the baseline methods in most cases.
Our main contributions are summarized as follows:
\begin{itemize}
    \item We develop PathNN, a neural network that computes node representations by aggregating paths of increasing length. 
    We derive three different variants of the model that focus on single shortest paths, all shortest paths and all simple paths, respectively.
    \item We prove that two of the variants are strictly more powerful than the WL algorithm and we empirically measure their expressive power.
    \item We evaluate the proposed model on several graph classification and regression datasets where it achieves performance comparable to state-of-the-art GNNs.
\end{itemize}

\section{Related Work}
\textbf{GNNs and kernels that process paths.}
Shortest path distances between nodes have been incorporated as structural features into several GNN architectures.
For instance, Graphormer encodes shortest path distances between two nodes as a bias term in the softmax attention~\cite{ying2021transformers}, while other models annotate nodes with features that emerge from shortest paths (\eg shortest path distances)~\cite{li2020distance,you2019position}.
On the other hand, PEGN uses shortest path distances to create persistence diagrams based on which messages between nodes are re-weighted~\cite{zhao2020persistence}.
Instead of aggregating the representations of each node's direct neighbors, some models such as SP-MPNN~\cite{abboud2022shortest} also consider the nodes at shortest path distance exactly $k$ from the node.
A recently proposed GNN framework, so-called Geodesic GNN~\cite{kong2022geodesic}, generates representations for pairs of nodes by aggregating the representations of the nodes on a single shortest path between the two nodes and also the direct neighbors of the two nodes that are on any of their shortest paths.
Perhaps the work most related to our approach is the PathNet model~\cite{sun2022beyond}, which also aggregates path information.
However, there are major differences between PathNet and our PathNNs.
PathNet samples paths instead of enumerating all of them, it is only evaluated on node classification datasets, and lastly, the authors do not provide an extensive study of the expressive power of the model.
There exist also models that instead of paths simulate walks which are then either aggregated~\cite{jin2022raw} or processed by a convolution neural network~\cite{toenshoff2021graph}.

Our proposed PathNNs are also related to graph kernels which compare paths of two graphs to each other.
Such kernels include the shortest path kernel~\cite{borgwardt2005shortest} and the GraphHopper kernel~\cite{feragen2013scalable}.
The former just compares shortest path distances to each other, while the latter is more similar to the proposed model since it also takes into account the attributes of the nodes that appear on a shortest path.

\begin{figure*}[t]
  \centering
  \subfigure[]{\includegraphics[width=.16\linewidth]{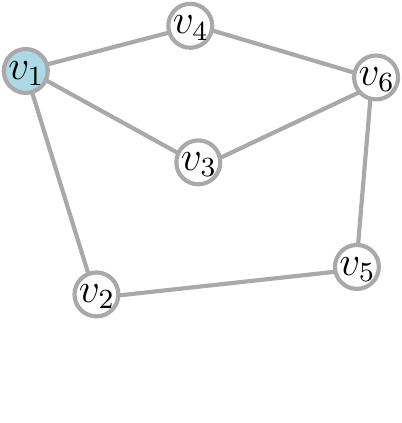}
  \label{fig:graph}}%
  \hfil
  \subfigure[]{\includegraphics[width=.12\linewidth]{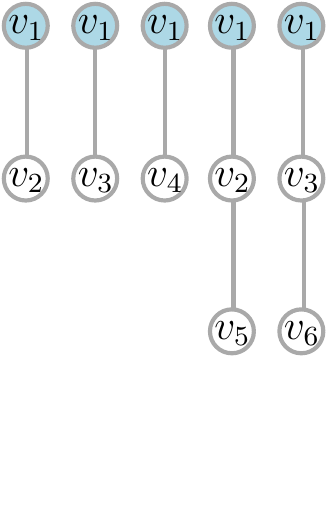}
  \label{fig:sp}}%
  \hfil
  \subfigure[]{\includegraphics[width=.146\linewidth]{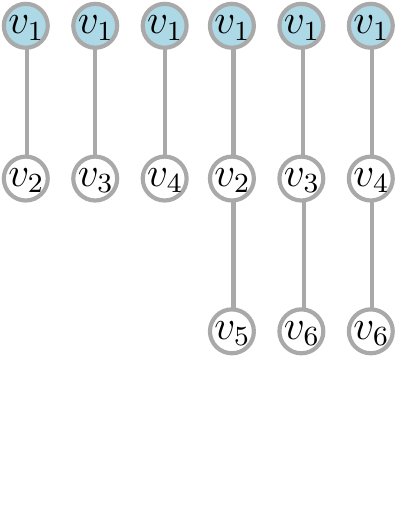}
  \label{fig:all_sp}}%
  \hfil
  \subfigure[]{\includegraphics[width=.276\linewidth]{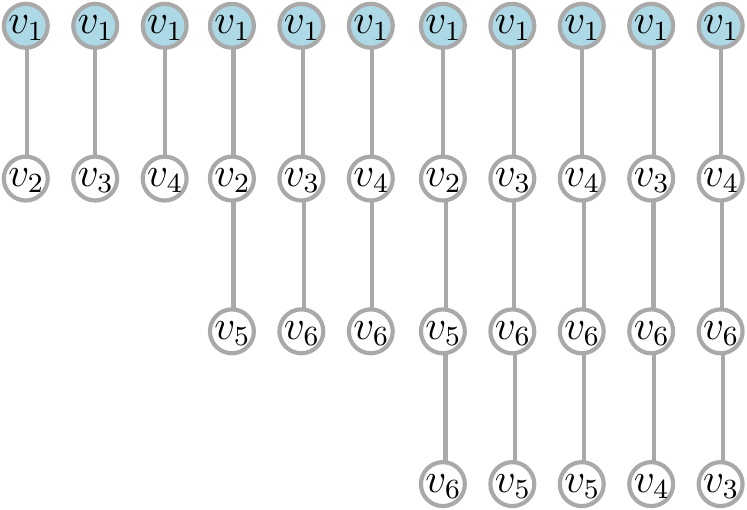}
  \label{fig:all_p}}%
  \caption{(a) A graph G with $n=6$ vertices. (b) Shortest paths of length up to $2$ starting from vertex $v_1$, (c) all shortest paths of length up to $2$ starting from vertex $v_1$ and (d) all paths of length up to $3$ starting from vertex $v_1$.}
  \label{fig:paths}
\end{figure*}

\textbf{Expressive power of GNNs.}
Since we will study the expressive power of our PathNNs, our work is furthermore related to the extensive literature exploring the expressive power of GNNs.
Several of those studies have investigated the connections between GNNs and the WL test of isomorphism~\cite{barcelo2020logical,geerts2022expressiveness}.
For instance, standard GNNs were shown to be at most as powerful as the WL algorithm in terms of distinguishing non-isomorphic graphs~\cite{morris2019weisfeiler,xu2019powerful}.
Other studies capitalized on high-order variants of the WL algorithm to derive new models that are more powerful than standard GNNs~\cite{morris2019weisfeiler,morris2020weisfeiler}.
Another line of research focused on $k$-order graph networks~\cite{maron2019invariant}.
Importantly, $k$-order graph networks were found to be at least as powerful as the $k$-WL graph isomorphism test in terms of distinguishing non-isomorphic graphs, while a reduced $2$-order network containing just a scaled identity operator, augmented with a single quadratic operation was shown to have $3$-WL discrimination power~\cite{maron2019provably}.
Furthermore, it was shown that a GNN with a maximum tensor order $2$ defined on the ring of equivariant functions can distinguish some pairs of non-isomorphic regular graphs with the same degree~\cite{chen2019equivalence}.
Graph isomorphism testing and invariant function approximation, the two main perspectives for studying the expressive power of GNNs, have been shown to be equivalent to each other~\cite{chen2019equivalence}.
Recently, a large body of work focused on making GNNs more powerful than WL, \eg by encoding vertex identifiers~\cite{vignac2020building}, taking into account all possible node permutations~\cite{murphy2019relational,dasoulas2020coloring}, using random features~\cite{sato2021random,abboud2021surprising}, using node features~\cite{you2021identity}, spectral information~\cite{balcilar2021breaking}, simplicial and cellular complexes~\cite{bodnar2021weisfeiler,bodnar2021weisfeiler2} and directional information~\cite{beaini2021directional}.
Furthermore, several recent studies extract and process subgraphs to make GNNs more expressive~\cite{nikolentzos2020k,bevilacqua2022equivariant}.
For example, the expressive power of GNNs can be increased by aggregating the representations of subgraphs (produced by standard GNNs) that arise from removing one or more vertices from a given graph~\cite{cotta2021reconstruction,papp2021dropgnn}.
Finally, it was recently shown that models that process each node's $k$-hop neighborhood and which aggregate nodes at shortest path distance exactly $k$ from that node (such as the SP-MPNN model~\cite{abboud2022shortest}) are more powerful than standard GNNs~\cite{feng2022powerful}, while the expressive power of those models can be further improved by taking into account the edges that connect the nodes at shortest path distance exactly $k$ from the considered node.

\begin{figure*}[t]
  \centering
  \subfigure[]{\includegraphics[width=.34\linewidth]{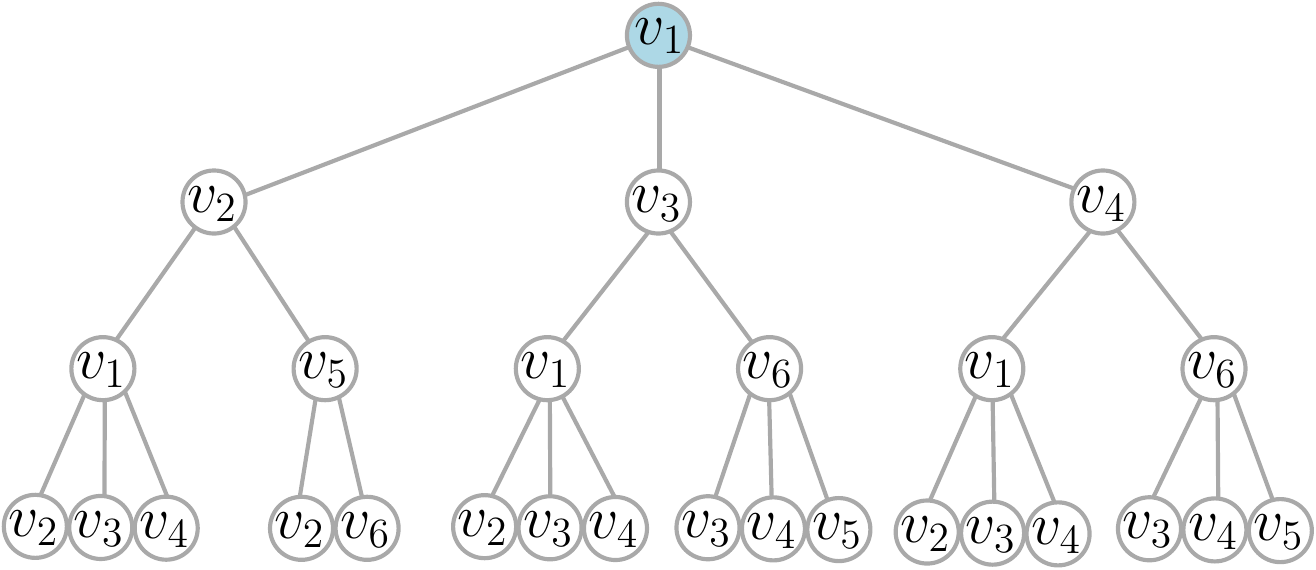}\label{fig:wl_tree}}%
  \hfil
  \subfigure[]{\includegraphics[width=.12\linewidth]{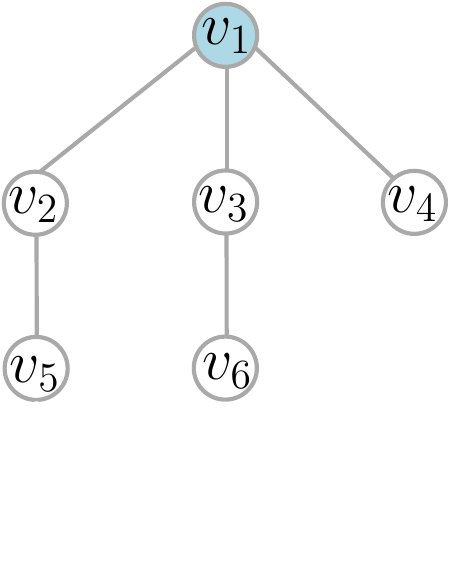}
  \label{fig:sp_tree}}%
  \hfil
  \subfigure[]{\includegraphics[width=.12\linewidth]{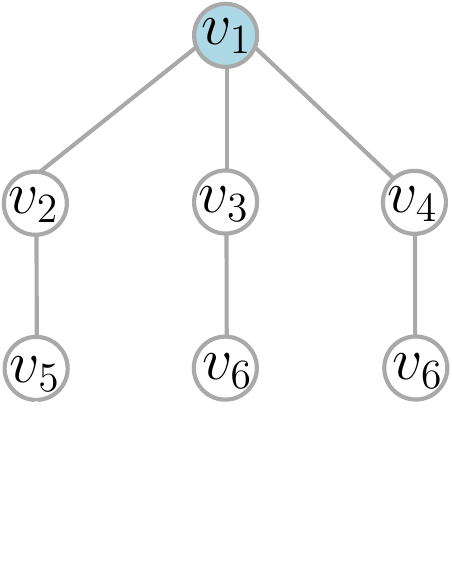}
  \label{fig:asp_tree}}%
  \hfil
  \subfigure[]{\includegraphics[width=.13\linewidth]{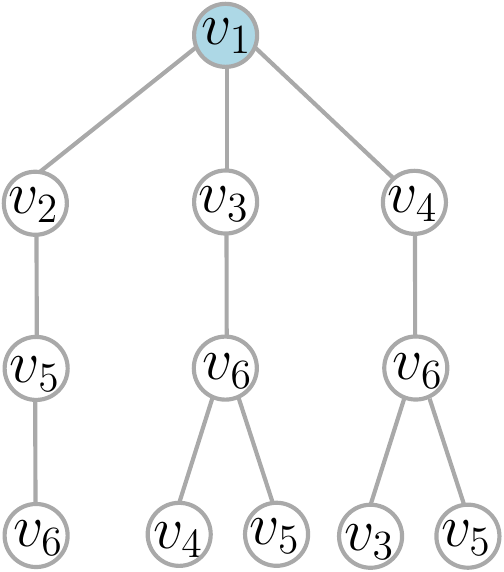}
  \label{fig:ap_tree}}%
  \caption{(a) WL-Tree rooted at node $v_1$ for the graph of Figure~\ref{fig:graph}. (b) One of the two $\mathcal{SP}$-Trees rooted at $v_1$ of height $2$. (c) $\mathcal{SP}^+$-Tree rooted at $v_1$ of height $2$ and (d) $\mathcal{AP}$-Tree rooted at $v_1$ of height $3$. Note that we consider only $\mathcal{SP}$ and $\mathcal{SP}^+$-Trees of height up to $2$ because there is no shortest path of length higher than $2$ that starts at $v_1$ in the input graph.}
  \label{fig:trees}
\end{figure*}

\section{Path Neural Networks}
In this section, we begin by introducing some key notation for graphs which will be used later, and then we describe Path Neural Networks (PathNNs), a message-passing GNN that learns representations of paths of increasing length which are aggregated to update node representations.
We first introduce the theoretical framework of PathNNs and then detail their architecture. 

\subsection{Notation}
Let $\oms \cms$ denote a multiset, \ie a generalized concept of a set that allows multiple instances for its elements. 
Let $G= (V,E)$ be an undirected graph consisting of a set of nodes $V$ and a set of edges $E \subseteq V \times V$. We denote by $n$ the number of nodes in $G$ and by $m$ its number of edges. The set $\mathcal{N}(v)$ represents the neighbors of vertex $v$. When considering attributed graphs, each vertex $v \in V$ is endowed with an initial node feature vector denoted by $\mathbf{x}_v$ that can contain categorical or real-valued properties of $v$.

A path from source node $v$ to target node $u$ is denoted by $\pi = [v_1, v_2, \dots, v_k]$ such that $v_1 =v$, $v_k =u$ and $(v_i, v_{i+1}) \in E$ for $i \in\{ 1,\dots,k-1\}.$ Note that paths only contain distinct vertices.  We denote by $\pi(j) = v_j$ the $j$-th node encountered when hopping along the path, and by $|\pi| = k-1$ its length, defined by the number of edges it contains. In this work, we consider various collections of paths. The first collection is the set of shortest paths denoted as $\mathcal{SP}$ containing a single shortest path for all possible node pair combinations. Multiple shortest paths between a source node and a target node might exist. We thus also consider the collection of all shortest paths that we denote as $\mathcal{SP}^+$, containing all possible shortest paths between every node pair combinations. 
A simple path can be any path, not necessarily the shortest, from a source node to a target node. We denote by $\mathcal{AP}$ the collection of all simple paths between node pair combinations. 
Figure~\ref{fig:paths} provides an example of these three collections of paths. Note that, we have $\mathcal{SP} \subseteq \mathcal{SP}^+ \subseteq \mathcal{AP}$. In practice, we only consider paths up to a fixed length $k$. We use the notation $\mathcal{P}_{v}^k = \{ \pi \in \mathcal{P}:\; \pi(1) = v,\; |\pi| = k \}$ for $\mathcal{P} \in \{\mathcal{SP}, \mathcal{SP}^+, \mathcal{AP}\}$ to denote all paths of length $k$ contained in $\mathcal{P}$ starting from node $v$. Similarly, $\mathcal{P}_{v}$ denotes the set of all paths in $\mathcal{P} \in \{\mathcal{SP}, \mathcal{SP}^+, \mathcal{AP}\}$ starting from node $v$ of any length.

\subsection{Theoretical Framework}
\label{sec:theory}
In this section, we assume that for any graph $G$, we have the collections of all paths of maximum length $K$ at our disposal for all nodes in $G$. We furthermore consider any collection of paths $\mathcal{P} \in \{\mathcal{SP}, \mathcal{SP}^+, \mathcal{AP}\}$ that is induced by a graph $G$.
All proofs can be found in Appendix~\ref{appendix:proofs}.
We begin by introducing the concept of \textit{WL-Trees} and \textit{Path-Trees}. 
Path-Trees are an intuitive path-based, rather than walk-based, analogue to WL-Trees. 
We will then show that Path-Trees are not able to distinguish graphs at a node-level for all collections of paths, which will motivate us to instead define a model operating on annotated sets of paths, which we will show to disambiguate graphs at least as well as WL-Trees. 

WL-Trees of a given root node are constructed one level at a time. 
At each iteration, level $k+1$ of a WL-Tree is constructed by setting the children of any node at level $k$ to its direct neighbors.
In the context of the WL algorithm the color of a node $v$ at iteration $k$ of the 1-WL algorithm represents a tree structure of height $k$ rooted at $v.$
The WL-Tree of the graph in Figure~\ref{fig:graph} is provided in Figure~\ref{fig:wl_tree}.
Similarly, we define Path-Trees, where we make use of the notation $\mathcal{T}$ to denote all Path-Trees, and $\mathcal{T}^{k}$ to denote Path-Trees of height $k$. 

\begin{definition}(Path-Tree rooted at $v$)
\label{def1}
    Let $G = (V,E)$ and $\mathcal{P} \in \{ \mathcal{SP}, \mathcal{SP}^+, \mathcal{AP}\}$ be a collection of any type-specific paths in $G$.
    A Path-Tree $P_v \subseteq \mathcal{T},$ induced by $\mathcal{P}_v$ for $v \in V,$ is a tree such that the node set at level $k$ of the tree is equal to the multisets of nodes that occur at position $k$ in the paths in $\mathcal{P}_v^{k},$ i.e., $\oms u: \pi(k)=u \text{ for } \pi\in \mathcal{P}_v^{k}\cms.$
    Nodes at level $k$ and level $k+1$ in the tree are connected if and only if they occur in adjacent positions $k$ and $k+1$ in a given path in the set of paths $\mathcal{P}_v,$ i.e., $(x,y) \in P_v$ if $\pi(k) = x$ and $\pi(k+1) = y$ for any $\pi \in \mathcal{P}_v$ such that each node at level $k+1$ is connected only to a single node at level $k.$ 
    The height $k$ Path-Tree $P_v^{k} \subseteq \mathcal{T}^{k}$ rooted at $v$ corresponds to the Path-Tree of $v$ pruned from all levels $l > k$.
 
\end{definition}
Different collections of paths lead to different Path-Trees, as shown in Figure~\ref{fig:trees}. 
To give the reader a more practical understanding of Path-Trees we now explain how a $k$ Path-Tree can be constructed from any collection of paths $\mathcal{P} \in \{ \mathcal{SP}, \mathcal{SP}^+, \mathcal{AP}\}$.
We iteratively build the Path-Tree for a given source node $v$ one level $k \in \{1, \ldots, K\}$ at a time, where we make use of the subset $\mathcal{P}_v^{k}$ of $\mathcal{P}_v \subseteq \mathcal{P}$ to construct level $k$ of the Path-Tree. 
In the first iteration we add the root node $v$ to the Path-Tree. 
Then, at subsequent iterations we add the multiset of nodes at position $k$ in the paths in $\mathcal{P}_v^{k}$ to the Path-Tree.
We then iteratively connect these added nodes to the Path-Tree via edges, where a given node $u$ of the added nodes is connected to a single leaf $w$ in the existing Path-Tree such that the ordered set of ancestors of $w$ in the Path-Tree is identical to the ordered set of nodes preceding $u$ in its path of length $k$ from which the addition of $u$ resulted. 

Intuitively, for a given graph $G$ and a given node $v$ the $\mathcal{SP}$-Tree rooted at $v$ is a subgraph of the $\mathcal{SP}^+$-Tree which is itself a subgraph of the $\mathcal{AP}$-Tree.
Another interesting property of Path-Trees is provided in the following lemma. 
\begin{lemma}
\label{lemma1}
    Let $G = (V,E)$ and $\mathcal{P} \in \{ \mathcal{SP}, \mathcal{SP}^+, \mathcal{AP}\}$ be any collection of paths in $G$. Let $P_v^{k}$ be the Path-Tree rooted at $v$ of height $k$ and $W_v^{k}$ be the WL-Tree rooted at $v$ of height $k$ for $v\in V$. Then $P_v^{k}$ is a subgraph of $W_v^{k}$.  
\end{lemma}
Lemma~\ref{lemma1} can be easily proved by noting that a WL-Tree rooted at node $v$ of height $k$ is equivalent to an enumeration of all walks that start at $v$ of length up to $k$.
Since paths are a special case of walks, Path-Trees are subgraphs of WL-Trees, for all collections of paths. 

Each type of Path-Trees contain a different type of structural information over the input graph.
$\mathcal{SP}$ and $\mathcal{SP}^+$ Trees contain the most compressed information over shortest paths.
In particular, they can only answer the question of \textit{what is the shortest path distance between a given node and any other node in $G$?} 
The $\mathcal{AP}$-Trees encode the largest amount of structural information over the input graph. 
Besides, they are not height-limited by the graph's diameter.
However, they usually grow exponentially with the height $k$ and the density of the input graph.
We now state our main theorem on the relation between $\mathcal{AP}$-Trees and WL-Trees.
\begin{theorem}
\label{theorem:simple_paths_wl}
    Let $G=(V,E)$ and $\mathcal{AP}$ be the collection of all simple paths in $G$.
    Let $\{ AP_v^{k},  AP_u^{k} \} \subseteq \mathcal{T}^{k}$ be the $\mathcal{AP}$-Trees of height $k$ rooted at $v$ and $u$, respectively, and $W_v^{k}$, $W_u^{k}$ be the WL-Trees of height $k$ rooted at $v$ and $u$, respectively, for $v, u \in V$. If $W_v^{k}$ is structurally different (\ie not isomorphic) than $W_u^{k}$, then $AP_v^k$ is structurally different than $AP_u^{k}$.
    Additionally, $AP_v^{k}$ and $AP_u^{k}$ can be different even if $W_v^{k}$ and $W_u^{k}$ are identical.  
\end{theorem}

Theorem~\ref{theorem:simple_paths_wl} states that, if at iteration $k$ 1-WL decides that two nodes have different colors, then their $\mathcal{AP}$-Trees are structurally different.
Additionally, it states that $\mathcal{AP}$-Trees are also able to disambiguate nodes that the 1-WL algorithm would determine to be structurally similar.
Hence, a model $f : \mathcal{AP} \xrightarrow[]{} \mathbb{R}^d$ that embeds $\mathcal{AP}$-Trees into $d$-dimensional vectors, such that non-isomorphic trees are embedded into different vectors, also equipped with an injective readout function is more expressive than the 1-WL algorithm.
Note that only considering $\mathcal{AP}$-Trees up to a fixed height $K$ is sufficient to distinguish a broad class of graphs. 

Now that we have considered the case of all paths $\mathcal{AP},$ we turn our attention to the case of the shortest paths in $\mathcal{SP}$ and $\mathcal{SP}^+.$
In Proposition \ref{prop:SP_example} we show that $\mathcal{SP}$-Trees and $\mathcal{SP}^+$-Trees are unable to capture all differences in WL-Trees. 

\begin{proposition}
    \label{prop:SP_example}
    Let $G=(V,E)$ and $\mathcal{P}$ be the collection of either single shortest paths $\mathcal{SP}$ or all shortest paths $\mathcal{SP}^+$ in $G$. Let $\{ P_v^{k},  P_u^{k} \} \subseteq \mathcal{T}^{k}$ be the $\mathcal{P}$-Trees of height $k$ rooted at $v$ and $u$, respectively, and $W_v^{k}$, $W_u^{k}$ be the WL-Trees of height $k$ rooted at $v$ and $u,$ respectively, for $v, u \in V$. If $W_v^{k}$ is structurally different than $W_u^{k}$, then $P_v^{k}$ is not necessarily structurally different from $P_u^{k}$.
\end{proposition}

Hence, Path-Trees based on the collections of shortest paths do not necessarily allow us to disambiguate individual nodes structurally even when the WL test does so. We therefore propose to operate on annotated sets of paths instead of Path-Trees. We use $\mathcal{\tilde{SP}}, \mathcal{\tilde{SP}}^+$ and $ \mathcal{\tilde{AP}}$ to denote annotated paths in the single shortest path, all shortest path and all simple path collections, respectively. The annotations of nodes in $\mathcal{\tilde{P}} \in \{\mathcal{\tilde{SP}}, \mathcal{\tilde{SP}}^+,\mathcal{\tilde{AP}}\}$ depend on the length of the considered path. For paths of length 1, i.e., $\tilde{P}^1 \in \mathcal{\tilde{P}},$ all nodes have equal annotations. For paths of length 2, i.e., $\tilde{P}^2 \in \mathcal{\tilde{P}},$ all nodes $v$ in these paths are annotated with hashes of their respective annotated path sets of length 1, i.e., $\tilde{P}^1_v.$ In general for paths of length $k,$ i.e., $\tilde{P}^k \in \mathcal{\tilde{P}},$ all nodes $v$ in these paths are annotated with hashes of their respective annotated path sets of length $k-1,$ i.e., $\tilde{P}^{k-1}_v.$ Each annotation of paths of length $k>2,$ is therefore a multiset of sequences of annotations. In Theorem \ref{thm:ExpressivityOfAnnotatedSetsOfPaths} we demonstrate that any annotated path set $\mathcal{\tilde{P}} \in \{\mathcal{\tilde{SP}}, \mathcal{\tilde{SP}}^+,\mathcal{\tilde{AP}}\}$ allows us to disambiguate individual nodes structurally at least as well as the WL test. Consequently, any model composed of injective functions that operates on annotated sets of paths, which notably includes the PathNNs that we will define in Section \ref{sec:ModelArchitecture}, is able to disambiguate graphs at least as well as the WL algorithm.

\begin{theorem} \label{thm:ExpressivityOfAnnotatedSetsOfPaths}
Let $G=(V,E)$ and $\mathcal{\tilde{P}} \in \{\mathcal{\tilde{SP}}, \mathcal{\tilde{SP}}^+,\mathcal{\tilde{AP}}\}$ denote an annotated set of paths, in which every node is annotated according to a recursive annotation scheme, in which a node $v$ occurring in a path of length $k$ is annotated by the hash of $v$'s annotated path sets of length $k-1,$ i.e., $\tilde{P}^{k-1}_v$ and the initial annotations of all nodes are equal. Let $\tilde{P}_v^{k}$ and $\tilde{P}_u^{k} $ be the annotated sets of paths of length $k$ emanating from nodes $v, u\in V$, respectively, and $ W_v^{k}$, $ W_u^{k}$ be the WL-Trees of height $k$ rooted at $v$ and $u,$ respectively, for $v, u \in V$. 

Then, if $ W_v^k$ and $ W_u^k$ are unequal, then $ \tilde{P}_v^k$ and $ \tilde{P}_u^k$ are unequal.
Additionally, $ \tilde{P}_v^k$ and $ \tilde{P}_v^k$ can be different even if $ W_v^k$ and $ W_v^k$ are identical.
\end{theorem}

We want to remark that depending on the particular shortest path that is sampled in the single shortest path collection $\mathcal{\tilde{SP}}$ it is possible that even isomorphic nodes have different annotated sets of paths. For $\mathcal{\tilde{SP}}^+$ and $\mathcal{\tilde{AP}}$ isomorphic nodes always have corresponding identical annotated sets of paths. Therefore, models operating on $\mathcal{\tilde{SP}}^+$ and $\mathcal{\tilde{AP}}$ are strictly more powerful than the WL algorithm for graph isomorphism. While models operating on $\mathcal{\tilde{SP}}$ can only disambiguate graphs at least as well as the WL algorithm and are not strictly more powerful, since also isomorphic graphs could be mapped to different representations.

\subsection{Model Architecture} \label{sec:ModelArchitecture}

\begin{figure*}[t!]
    \centering
    \subfigure[First iteration of PathNN-$\mathcal{SP}$.]{\includegraphics[width=0.6\linewidth]{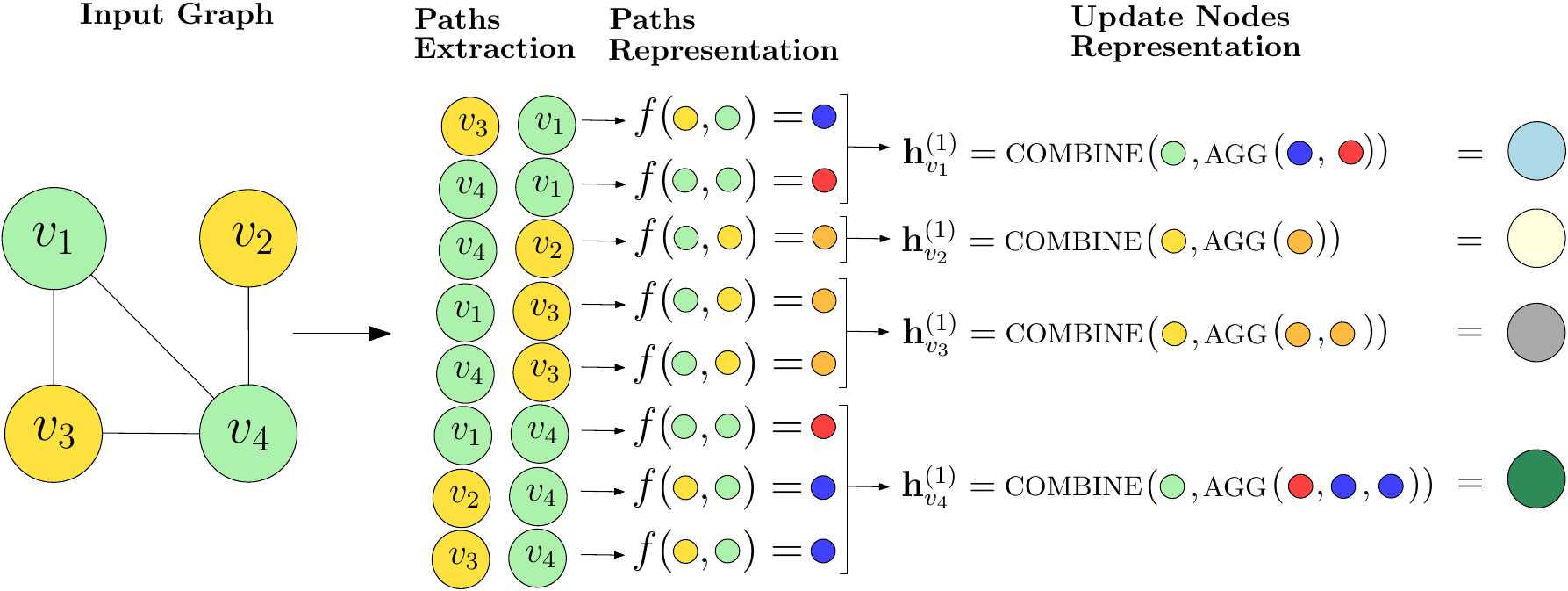}}
    \vfill
    \subfigure[Second iteration of PathNN-$\mathcal{SP}$. ]{\includegraphics[width=0.6\linewidth]{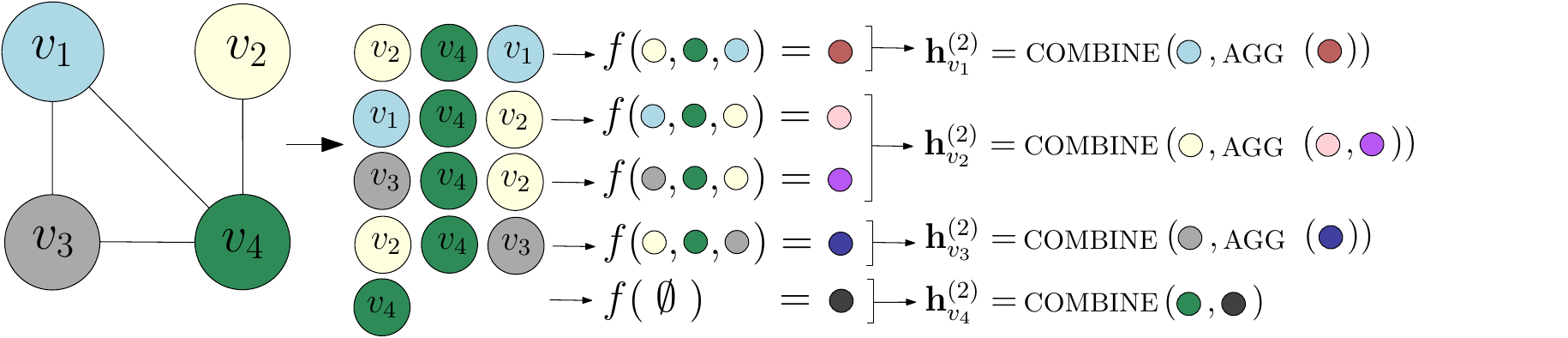}}
    \caption{Aggregation process of PathNN with $\mathcal{P} = \mathcal{SP}$ for two iterations. Node colors corresponds to node feature vectors.}
    \label{fig:pathnn}
\end{figure*}

We adopt a message passing scheme that iteratively updates node representations using paths of increasing length.
This message passing scheme allows us to refine sets of paths with additional structural information included in paths of length smaller than $K$.

Noting that paths are ordered sequences of nodes, PathNNs learn to embed paths into $d$-dimensional feature vectors, using a function operating on sequences $f : \mathbb{R}^{k \times d} \rightarrow \mathbb{R}^d,$ where $k$ is the path length.
PathNNs employ a message-passing scheme that aggregates path embeddings to form updated node representations.
Specifically, at each layer $k,$ paths of length $k$ are embedded by $f$ and aggregated.
Let $\pi$ be any path of length $k$.
Then, PathNNs compute the embedding of $\pi$ by using the node representations of layer~$k-1$ 
\begin{equation}
    \mathbf{h}_\pi^{(k)} = f \left( \left[\mathbf{h}_{\pi(k+1)}^{(k-1)}, \dots, \mathbf{h}_{\pi(1)}^{(k-1)} \right]\right).
\end{equation}
PathNNs start to encode paths of length $1$ in the first layer (\ie sequences of a node and its neighbors' representations), then iteratively encode paths of increasing length.
Path representations form the message in the message passing architecture and are used to update node feature vectors in a similar way to traditional MPNNs 
\begin{align*}
    \mathbf{a}_v^{(k)} &= \text{AGGREGATE}^{(k)} \left( \Bigoms \mathbf{h}_\pi^{(k)} | \pi \in \mathcal{P}_v^{k} \Bigcms \right),\\
    \mathbf{h}_v^{(k)} &= \text{COMBINE}^{(k)} \left( \mathbf{h}_v^{(k-1)}, \, \mathbf{a}_v^{(k)}\right).
\end{align*} 
Paths of length up to $K$ are given as input to PathNNs by stacking $K$ layers.
Figure~\ref{fig:pathnn} illustrates the functioning of PathNNs. 
Graph representations are obtained using a permutation-invariant readout function
\begin{equation*}
    \mathbf{h}_G = \text{READOUT} \left( \Bigoms \mathbf{h}_v^{(K)} | v \in V \Bigcms \right).
\end{equation*}
Without loss of generality, we propose to use a simple instance of PathNNs in this paper.
The function $f$ operating on sequences is modeled and learned by a Long-Short-Term-Memory (LSTM) cell~\cite{hochreiter1997long}.
LSTMs are a valid choice for the function $f$ thanks to the universal approximation results for Recurrent Neural Network~\cite{hammer2000approximation}: they can approximate any injective sequence function arbitrarily well in probability, which is a desirable and necessary property for $f$.
Note that the last element in the sequence is the starting node representation.
The LSTM operates on reversed paths as we hypothesize that the most important information in the sequence is the starting node's current representation.
We project node initial feature vectors to match the LSTM's hidden dimension with a 2-layer Multilayer Perceptron (MLP).
Using paths embeddings, PathNNs then update node representations using the following rule 
\begin{align}
\label{update_function}
	\mathbf{g}_v^{(k)} &= \text{\small NORM}^{(k)} \left( \mathbf{h}_v^{(k-1)} + \sum_{\pi \in \mathcal{P}_v^k} \mathbf{h}_\pi^{(k)} \right), \\
\label{update_function_2}
    \mathbf{h}_v^{(k)} &= \text{\large$ \phi $} \left( \mathbf{g}_v^{(k)}  \right).
\end{align}
Noting that the collection of paths $\mathcal{P}_v^k$ grows larger with graph density, the right hand side of Equation~(\ref{update_function}) can be of very high magnitude, leading to numerical instabilities. 
A Batch Normalization (BN) layer~\cite{ioffe2015batch} is thus applied in Equation~(\ref{update_function}) to avoid such situations during training. 
After normalization, node representations are passed through a $\phi$ function, which can be either the identity function or a 2-layer MLP. 
Both functions are a valid choice as sum and 2-layer MLPs are injective functions over multisets \cite{xu2019powerful}. 
Equipped with the identity aggregation function, the number of parameters of PathNNs only slightly increases with path length (caused by BN parameters), resulting in a low number of trainable parameters even for higher path lengths. 
Replacing the identity function with an MLP allows finer updated node representations but leads to higher time and memory complexity. 
Finally, PathNNs use the sum over the multiset of final node representations as a readout function to produce a vector representation of the entire graph 
\begin{equation}
\label{readout_function}
    \mathbf{h}_G = \sum\limits_{v \in V} \mathbf{h}_v^{(K)}.
\end{equation}

\subsection{Time and Space Complexity}
To enumerate all shortest paths and all simple paths of length up to $k$ from a source vertex to all other vertices, we use the depth-first search (DFS) algorithm.
The time complexity of the algorithm is at most $\mathcal{O}(b^k)$, where $b$ is the branching factor of the graph which is upper bounded by the maximum of the nodes' degrees.
Thus, for all the nodes of the graph, the time complexity is $\mathcal{O}(nb^k)$.
The space complexity is $\mathcal{O}(nbk)$ if duplicate nodes are not eliminated.

Real-world graphs are often small (\eg molecules) and/or sparse (\eg social networks), i.e., $n$ is often small and/or $b \ll n$.
Thus, for bounded $k$, the time and space complexity for enumerating the paths is not prohibitive.

The running time of the model depends on the number of paths which is $\mathcal{O}(nb^k)$.
Typically, the larger the value of $k$, the larger the number of paths and therefore, the complexity of our model increases as a function of $k$.
For $k=1$, the complexity of our model is comparable to that of standard GNNs which aggregate $2m \approx nb$ representations.
However, for larger values of $k$, the time complexity of our model is greater than that of standard GNNs which only aggregate $2m$ representations in all neighborhood aggregation layers.
We report the empirical running times of our PathNNs on two real-world datasets in Appendix~\ref{sec:running_time}.

\section{Experimental Evaluation}
We now evaluate the performance of our PathNNs in synthetic experiments specifically designed to exhibit the expressiveness of GNNs in Section~\ref{sec:SyntheticDatasets} and on a range of real-world datasets in Section~\ref{sec:RealWorldDatasets}. 

\subsection{Synthetic Datasets}
\label{sec:SyntheticDatasets}

\textbf{Datasets.}
We use $3$ publicly available datasets: ($1$) the Circular Skip Link (CSL) dataset~\cite{murphy2019relational}; ($2$) the EXP dataset~\cite{abboud2021surprising}; ($3$) the SR dataset.

\textbf{Experimental setup.}
For all experiments, the aggregation function $\phi$ of Equation~\eqref{update_function_2} is set to the identity function and the normalization layer of Equation~\eqref{update_function} is removed. 
We set initial node features to be vectors of ones, and process them using a $2$-layer MLP.
A 1-layer MLP is applied to the final graph representation to generate predictions. 

The benchmarks CSL and EXP-Class require the classification of graphs into isomorphism classes. 
To evaluate the model's performance, we used $5$-fold cross validation on CSL and $4$-fold cross validation on EXP-Class. 
Graphs contained in the CSL dataset present a maximum diameter of $11$. 
We thus set $K$ to $11$ for PathNN-$\mathcal{SP}$ and PathNN-$\mathcal{SP}^+$.  
$K$ is set to $5$ for PathNN-$\mathcal{AP}$.  
As EXP graphs are disconnected, we empirically set $K$ to $10$ for PathNN-$\mathcal{SP}$ and PathNN-$\mathcal{SP}^+$ and $5$ for PathNN-$\mathcal{AP}$ since these values allow the models to achieve perfect performance in this task. 
For both experiments and all models, we train for $200$ epochs using the Adam optimizer with learning rate $10^{-3}$.
The hidden dimension size is set to $64$. 
Batch sizes are set to $32$ except for PathNN-$\mathcal{AP}$ where we set it to $8$ for CSL and $16$ for CEXP to be able to fit all paths in memory.
For PathNN-$\mathcal{AP}$, we apply Euclidean normalization before feeding the represenentations to the LSTM.
Euclidean normalization is used instead of BN because the latter results in training instabilities.
All PathNN-$\mathcal{AP}$ models are trained with distance encoding.

For the SR and EXP-Iso datasets, we investigate whether the proposed models have the right inductive bias to distinguish these pairs of graphs. 
Similarly to~\citet{bodnar2021weisfeiler2}, we consider two graphs isomorphic if the Euclidean distance between their representation is below a fixed threshold $\varepsilon$. 
Graph representations are computed by an untrained model variant where the prediction layer is removed. 
We remove the normalization layer of Equation~\eqref{update_function} and instead apply Euclidean normalization to the LSTM's input.
For EXP-Iso, we use the same $K$ as in the experiments on EXP-Class.
For the SR datasets, we set $K$ to $4$ as retrieving all paths of lengths higher than $4$ is computationally challenging on these densely connected graphs.  
Hidden dimension size is set to $16$ and $\varepsilon$ to $10^{-5}$.
The experiment is repeated with $5$ different seeds.

\begin{table}[t]
\centering
\caption{Test set classification accuracy (CSL, EXP-Class) and number of undistinguished pairs of graphs (EXP-Iso). Best results are highlighted in bold.}
\label{tab:exp_results}
\renewcommand{\arraystretch}{1.2}
\resizebox{\columnwidth}{!}{%
\begin{tabular}{lccc}
\toprule
Model & \textbf{CSL} $\uparrow$ & \textbf{EXP-Class} $\uparrow$ & \textbf{EXP-Iso} $\downarrow$ \\
\midrule
GIN~\cite{xu2019powerful} & \phantom{0}10.0 $\pm$ 0.0\phantom{0} & \phantom{0}50.0 $\pm$ 0.0  & 600 \\ 
3WLGNN~\cite{maron2019provably} & \phantom{0}97.8 $\pm$ 10.9 & \textbf{100.0} $\pm$ 0.0 & \textbf{0}\\
\midrule
PathNN-$\mathcal{SP}$ & \phantom{0}90.0 $\pm$ 0.0\phantom{0} & \textbf{100.0} $\pm$ 0.0 &\textbf{0}\\
PathNN-$\mathcal{SP}^+$ & \textbf{100.0} $\pm$ 0.0\phantom{0} & \textbf{100.0} $\pm$ 0.0 & \textbf{0} \\
PathNN-$\mathcal{AP}$ & \textbf{100.0} $\pm$ 0.0\phantom{0} & \textbf{100.0} $\pm$ 0.0 & \textbf{0} \\
\bottomrule
\end{tabular}
}
\end{table}

\begin{figure}
\centering
    \includegraphics[width=\linewidth]{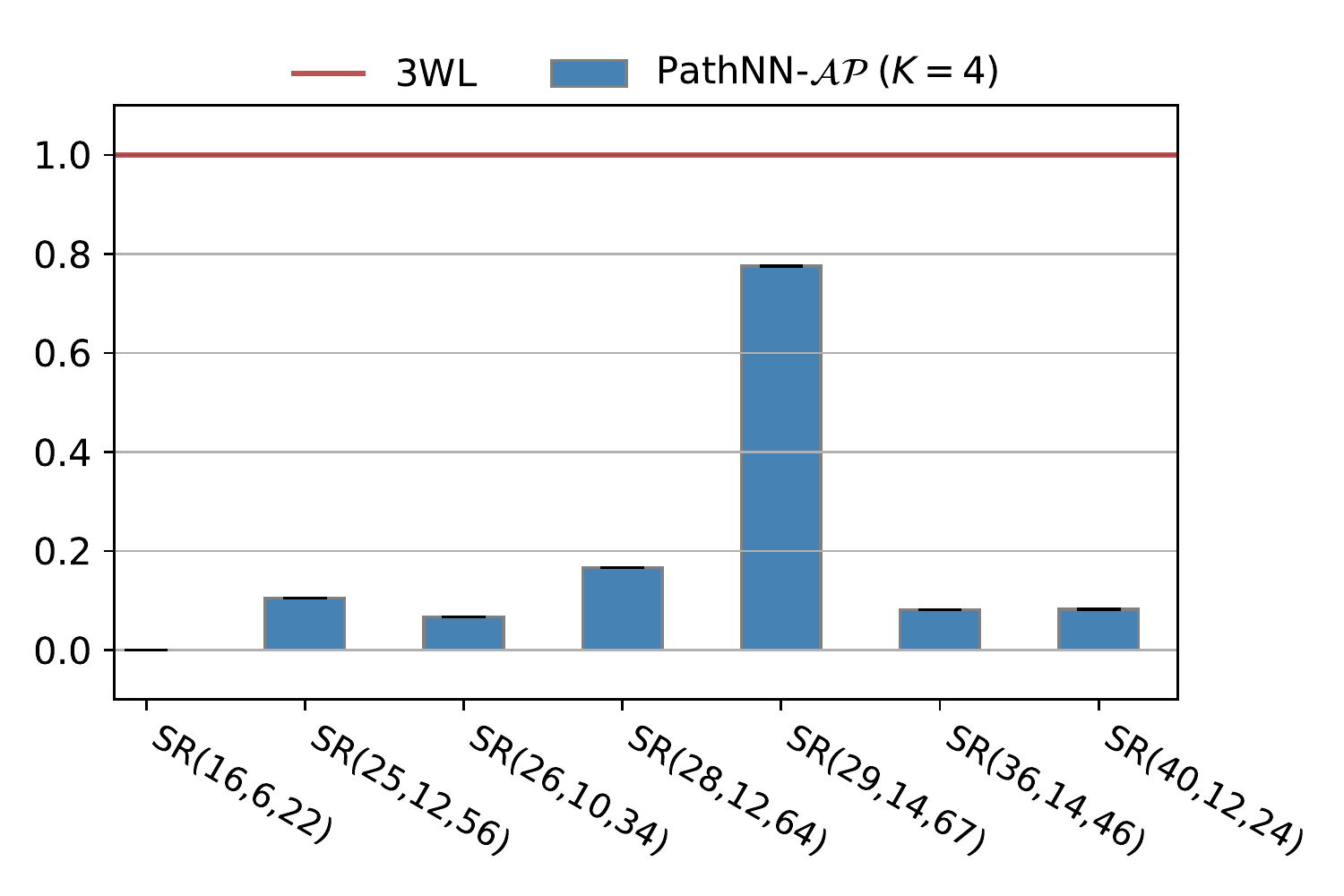}
    \vspace{-.5cm}
    \caption{Failure rate on different instances of the SR dataset.}
    \label{fig:sr}
\end{figure}

\begin{table*}[t]
\centering
\caption{Classification accuracy ($\pm$ standard deviation) of our PathNNs and the baselines on the datasets from the TUDataset collection. Best performance is highlighted in \textbf{bold}. 
OOM means out-of-memory and NA means not available.}
\label{tab:results_tudatasets}
\scriptsize
\renewcommand{\arraystretch}{1.2}
\begin{tabular}{l|cccccc}
\toprule
& \textbf{DD} & \textbf{PROTEINS}  & \textbf{NCI1} & \textbf{ENZYMES} & \textbf{IMDB-B} & \textbf{IMDB-M} \\
\midrule
DGCNN~\cite{zhang2018end} & 76.6 $\pm$ 4.5 & 72.9 $\pm$ 3.5 &  76.4 $\pm$ 1.7 &  38.9 $\pm$ 5.7  & 69.2 $\pm$ 3.0 & 45.6 $\pm$ 3.4\\ 
DiffPool~\cite{ying2018hierarchical} & 75.0 $\pm$ 4.3 & 73.7 $\pm$ 3.5 &  76.9 $\pm$ 1.9 & 59.5 $\pm$ 5.6 & 68.4 $\pm$ 3.3 & 45.6 $\pm$ 3.4\\ 
ECC~\cite{simonovsky2017dynamic} & 72.6 $\pm$ 4.1 & 72.3 $\pm$ 3.4 &  76.2 $\pm$ 1.4 &  29.5 $\pm$ 8.2 & 67.7 $\pm$ 2.8 & 43.5 $\pm$ 3.1\\ 
GIN~\cite{xu2019powerful} & 75.3 $\pm$ 2.9 & 73.3 $\pm$ 4.0 & 80.0 $\pm$ 1.4 &  59.6 $\pm$ 4.5 & 71.2 $\pm$ 3.9 & 48.5 $\pm$ 3.3 \\ 
GraphSAGE~\cite{hamilton2017inductive} & 72.9 $\pm$ 2.0 & 73.0 $\pm$ 4.5 &  76.0 $\pm$ 1.8  & 58.2 $\pm$ 6.0 & 68.8 $\pm$ 4.5 & 47.6 $\pm$ 3.5 \\
GAT~\cite{velivckovic2018graph} & 73.9 $\pm$ 3.4 & 70.9 $\pm$ 2.7 & 77.3 $\pm$ 2.5 & 49.5 $\pm$ 8.9 & 69.2 $\pm$ 4.8 & 48.2 $\pm$ 4.9 \\
SPN ($K=1$)~\cite{abboud2022shortest} & 72.7 $\pm$ 2.6 & 71.0 $\pm$ 3.7 & 80.0 $\pm$ 1.5 & 67.5 $\pm$ 5.5 & NA & NA \\ 
SPN ($K=5$)~\cite{abboud2022shortest} & 77.4 $\pm$ 3.8 & 74.2 $\pm$ 2.7 & 78.6 $\pm$ 3.8 & 69.4 $\pm$ 6.2 & NA & NA \\ 
PathNet ($N=10, K=2$)~\cite{sun2022beyond} & OOM & 70.5 $\pm$ 3.9 & 64.1 $\pm$ 2.3 & 69.3 $\pm$ 5.4 & 70.4 $\pm$ 3.8 & 49.1 $\pm$ 3.6   \\
Nested GNN~\cite{zhang2021nested} & \textbf{77.8} $\pm$ 3.9 & 74.2 $\pm$ 3.7 & NA & 31.2 $\pm$ 6.7 & NA & NA \\ 
\midrule
PathNN-$\mathcal{P\phantom{S}}\phantom{^+}$ $\, (K=1)$ & 76.9 $\pm$ 3.7 & \textbf{75.2} $\pm$ 3.9 & 77.5 $\pm$ 1.6 & \textbf{73.0} $\pm$ 5.2 & 
\textbf{72.6} $\pm$ 3.3 & \textbf{50.8} $\pm$ 4.5 \\
PathNN-$\mathcal{SP}\phantom{^+}$ $\, (K=2)$ & 75.3 $\pm$ 2.7 & 73.1 $\pm$ 3.1 & 82.0 $\pm$ 1.6 & 71.6 $\pm$ 6.4 &
70.8 $\pm$ 3.5 & 50.0 $\pm$ 4.1 \\
PathNN-$\mathcal{SP}\phantom{^+}$ $\, (K=3)$ & 77.0 $\pm$ 3.1 & 72.2 $\pm$ 2.7 & 82.2 $\pm$ 1.7 & 69.2 $\pm$ 4.7 & 
- & -\\
PathNN-$\mathcal{SP}^+$ $\, (K=2)$ & 74.7 $\pm$ 3.0 & 73.1 $\pm$ 3.7 & 81.0 $\pm$ 1.4 & 72.5 $\pm$ 5.3 &
70.5 $\pm$ 3.4 & 50.7 $\pm$ 4.5 \\
PathNN-$\mathcal{SP}^+$ $\, (K=3)$ & 76.5 $\pm$ 4.6 & 73.2 $\pm$ 3.3 &  \textbf{82.3} $\pm$ 1.9 & 70.4 $\pm$ 3.1  &
- & -\\
PathNN-$\mathcal{AP}\phantom{^*}$ $\, (K=2)$ & 75.0 $\pm$ 4.4 & 73.1 $\pm$ 4.9 & 81.3 $\pm$ 1.8 & 71.8 $\pm$ 4.8 &
71.7 $\pm$ 3.6 & 49.8 $\pm$ 4.2 \\
PathNN-$\mathcal{AP}\phantom{^*}$ $\, (K=3)$ & OOM & 73.1 $\pm$ 4.0& 82.3 $\pm$ 1.7 & 69.0 $\pm$ 5.3 &
OOM & OOM \\
\bottomrule
\end{tabular}
\end{table*}

\textbf{Results.}
The results are given in Table~\ref{tab:exp_results} and in Figure~\ref{fig:sr}.
We observe that the two most expressive variants of the proposed model, namely Path-$\mathcal{SP}^+$ and Path-$\mathcal{AP},$ can distinguish all $10$ isomorphism classes present in the CSL dataset.
On the other hand, PathNN-$\mathcal{SP}$ fails to distinguish one of the $10$ classes, but still significantly outperforms the GIN model.
All three variants of the proposed model achieve perfect accuracy on EXP-Class and never fail on the EXP-Iso dataset.
Those datasets contain graphs that are not distinguishable by $1$-WL, and thus, not surprisingly, GIN maps the graphs of each pair to the same vector.
Finally, we evaluate the most expressive variant of our model (\ie PathNN-$\mathcal{AP}$) on the SR dataset whose instances contain graphs not distinguishable by $3$-WL.
We can see in Figure~\ref{fig:sr} that in most cases, PathNN-$\mathcal{AP}$ can successfully distinguish more than $80\%$ of the graphs contained in each instance of the dataset.
SR($29,14,6,7$) seems to be particularly hard for PathNN-$\mathcal{AP}$ since its failure rate on this instance is high compared to its performance on other instances. 
Overall, our experiments confirm the high expressiveness of the proposed model in terms of distinguishing non-isomorphic graphs.

\subsection{Real-World Datasets}
\label{sec:RealWorldDatasets}

\textbf{Datasets.}
We evaluate the proposed model on $6$ datasets contained in the TUDataset collection~\cite{morris2020tudataset}: DD, NCI1, PROTEINS, ENZYMES, IMDB-B and IMDB-M.
We also evaluate the proposed model on ogbg-molhiv, a molecular property prediction dataset from the Open Graph Benchmark (OGB)~\cite{hu2020open}.
We conduct an experiment on the ZINC 12K dataset~\cite{dwivedi2020benchmarking}.
Finally, we experiment with Peptides-struct and Peptides-func~\cite{dwivedi2022long}, two datasets that require long-range dependencies between nodes to be captured.

\textbf{Experiment setup.}
In all experiments, we use 2-layer MLPs with BN to map initial node representations to the desired dimension, and 2-layer MLPs without BN for prediction. All PathNN-$\mathcal{AP}$ are trained with distance encoding. 

Following~\citet{errica2019fair}, we evaluate TUDatasets using a $10$-fold cross validation using their provided data splits. 
We let $\phi$ be a ReLU non-linearity and choose hidden dimension size from $\{ 32, 64\}$.
We apply dropout to the input of the LSTM layer and between the first and second layer of the final MLP. 
The dropout rate is chosen from $\{0, 0.5\}$. 
We run the experiment for various values of $K \in \{1,2,3\}$ to analyze the effect of increased path length on performance. 
The batch size is set to $32$ for all datasets and all values of $K$ except for simple paths on DD with $K = 2$ where we had to decrease batch size to $16$ because of memory constraints. 
The Adam optimizer is used with fixed learning rate $0.001$.
The diameter of the graphs contained in IMDB-B and IMDB-M is at most equal to $2$.
Thus, for those two datasets, we do not provide results for paths of length up to $K=3$.
Similarly to~\citet{errica2019fair}, we use one-hot encodings of given node attributes for all datasets except IMDB-B and IMDB-M, where we instead use one-hot encodings of node degrees. 
We also include the $18$ continuous node feature vectors available for ENZYMES. 
We fit each model for $500$ epochs and stop training if validation accuracy does not increase for $250$ epochs. 

All other experiments are conducted using available data splits. 
We set $\phi$ to a 2-layer MLP with BN for all experiments and set $K$ and the layer's hidden size to respect the $500K$ parameter budget for ZINC, Peptides-functional and Peptides-structural. 
Details about the hyperparameter configuration can be found in Appendix~\ref{appendix:hyperparameters}. 
All of these datasets contain additional edge features. 
Similarly to the Distance aware LSTM cell described in Appendix~\ref{appendix:distance_encoding}, we build an Edge LSTM cell that takes as input a sequence of node and edge representations. 
For $\mathcal{AP},$ the LSTM cell encodes both edges in the path and distance from the central node. 
Further detail can be found in Appendix~\ref{appendix:edge_encoding}.

\textbf{Results.}
Table~\ref{tab:results_tudatasets} illustrates the classification accuracy achieved by the proposed PathNNs and the baselines on the six datasets from the TUDataset collection.
We observe that our PathNNs outperform the baselines on $5$ out of the $6$ datasets.
When $K=1$, all three variants of our model are identical, since all path variants are identical when paths of length one are considered. 
We denote this variant by PathNN-$\mathcal{P}$. 
On most datasets, PathNN-$\mathcal{P}$ provides the highest accuracy.
This is not surprising for IMDB-B and IMDB-M since the graphs contained in these datasets are ego-networks of radius $2$.
On the other hand, it appears that on DD and NCI1, more global information needs to be captured.
On these two datasets, models that consider paths of length up to $K=3$ achieve the highest accuracy.
On some datasets, the proposed model significantly outperforms the baselines.
Notably, on the ENZYMES, NCI1 and IMDB-M datasets, our PathNNs offer respective absolute improvements of $3.6\%$, $2.3\%$ and $1.7\%$ in accuracy over the best competitor.
Overall, our results indicate that PathNNs achieve high levels of performance on the TUDatasets.

\begin{table}[t]
\centering
\caption{Results on the Peptides-Functional and Peptides-Structural datasets ($\pm$ standard deviation). Evaluation metrics are Average Precision and Mean Absolute Error, respectively. Best performance is highlighted in \textbf{bold}. Parameter budget is set to $500$K parameters. Results are averaged over $4$ random seeds.}
\label{tab:results_peptides}
\renewcommand{\arraystretch}{1.2}
\resizebox{\columnwidth}{!}{%
\begin{tabular}{l|cc|cc}
\toprule
& $K$ & \textbf{Peptides-Functional}  $\uparrow$ & $K$ & \textbf{Peptides-Structural} $\downarrow$ \\
\midrule
GCN~\cite{kipf2017semi} & 5 & 59.30 $\pm$ 0.23 & 5 & 0.3496 $\pm$ 0.0013 \\
GIN~\cite{xu2019powerful} & 5 & 54.98 $\pm$ 0.79 & 5 & 0.3547 $\pm$ 0.0045 \\
GatedGCN & \multirow{2}{*}{5} & \multirow{2}{*}{58.64 $\pm$ 0.77} & \multirow{2}{*}{5} & \multirow{2}{*}{0.3420 $\pm$ 0.0013}\\
\cite{bresson2017residual} & & & & \\
\midrule
PathNN-$\mathcal{SP}$ & 8 & \textbf{68.16} $\pm$ 0.26 & 4 & 0.2545 $\pm$ 0.0032\\
PathNN-$\mathcal{SP}^+$ & 8 & 67.84 $\pm$ 0.52 & 4 & \textbf{0.2540} $\pm$ 0.0046 \\
PathNN-$\mathcal{AP}$ & 7 & 68.07 $\pm$ 0.72 & 4 & 0.2569 $\pm$ 0.0030 \\
\bottomrule
\end{tabular}
}
\end{table}

We next evaluate the proposed model on the two datasets that require long-range interaction reasoning to achieve strong performance.
The results are shown in Table~\ref{tab:results_peptides}.
We can see that the variants of our PathNNs outperform the baselines on both datasets.
On Peptides-Functional, our model offers a respective absolute improvement of $8.86\%$ in average precision over GCN, while on Peptides-Structural, it offers a respective absolute improvement of $8.80\%$ in mean absolute error over GatedGCN.
To summarize, our results suggest that PathNNs can better capture long-range interactions between nodes than the baseline models.

\begin{table}[t]
\caption{ROC-AUC score ($\pm$ standard deviation) of the different methods on the ogbg-molhiv dataset. Results are averaged over $10$ random seeds. Best performance is highlighted in \textbf{bold}.}
\label{tab:results_ogb}
\centering
\scriptsize
\renewcommand{\arraystretch}{1.2}
\begin{tabular}{l|c}
\toprule
 & \textbf{ogbg-molhiv} $\uparrow$ \\
\midrule
GCN~\cite{kipf2017semi} & 76.06 $\pm$ 0.97 \\
GIN~\cite{xu2019powerful} & 75.58 $\pm$ 1.40 \\
GCN+FLAG~\cite{kong2020flag} & 76.83 $\pm$ 1.02 \\
GIN+FLAG~\cite{kong2020flag} & 76.54 $\pm$ 1.14 \\
GSN~\cite{bouritsas2022improving} & 77.99 $\pm$ 1.00 \\
HIMP~\cite{fey2020hierarchical} & 78.80 $\pm$ 0.82 \\
PNA~\cite{corso2020principal} & 79.05 $\pm$ 1.32 \\
DGN~\cite{beaini2021directional} & 79.70 $\pm$ 0.97 \\
Graphormer~\cite{ying2021transformers} & 80.51 $\pm$ 0.53 \\
CIN~\cite{bodnar2021weisfeiler2} & \textbf{80.94} $\pm$ 0.57 \\ 
ESAN~\cite{bevilacqua2022equivariant} & 78.00 $\pm$ 1.42 \\
E-SPN~\cite{abboud2022shortest} & 77.10 $\pm$ 1.20 \\
GRWNN~\cite{nikolentzos2023geometric} & 78.38 $\pm$ 0.99 \\
AgentNet~\cite{martinkus2023agent} & 78.33 $\pm$ 0.69 \\
\midrule
PathNN-$\mathcal{SP}\phantom{^+}$ $\, (K=2)$ & 78.62 $\pm$ 1.30 \\
PathNN-$\mathcal{SP}^+$ $\, (K=2)$ & 79.17 $\pm$ 1.09 \\
PathNN-$\mathcal{AP}\phantom{^*}$ $\, (K=2)$ & 78.84 $\pm$ 1.46 \\
\bottomrule
\end{tabular}
\end{table}

Table~\ref{tab:results_ogb} shows the ROC-AUC of the different methods on the ogbg-molhiv dataset.
We observe that PathNN is ranked as the third best model on this dataset.
Among the three variants of the proposed model, PathNN-$\mathcal{SP}^+$ achieves the highest ROC-AUC score.
The other two variants perform slightly worse than PathNN-$\mathcal{SP}^+$.
This experiment validates the effectiveness of the proposed model on large graph classification datasets.

\begin{table}[t]
\centering
\caption{Mean absolute error ($\pm$ standard deviation) of the different methods on the ZINC$12$K datasets. Results are averaged over $10$ random seeds. Best performance is highlighted in \textbf{bold}. Parameter budget is set to $500$K parameters.}
\label{tab:results_zinc}
\scriptsize
\renewcommand{\arraystretch}{1.2}
\begin{tabular}{l|cc}
\toprule
& $K$ & \textbf{ZINC12K} $\downarrow$ \\
\midrule
GCN~\cite{kipf2017semi} & 16 & 0.278 $\pm$ 0.003 \\
GraphSAGE~\cite{hamilton2017inductive} & 16 & 0.398 $\pm$ 0.002 \\
MoNet~\cite{monti2017geometric} & 16 & 0.292 $\pm$ 0.006 \\
GAT~\cite{velivckovic2018graph} & 16 & 0.384 $\pm$ 0.007 \\
GIN~\cite{xu2019powerful} & 5 & 0.387 $\pm$ 0.015 \\
GatedGCN~\cite{bresson2017residual} & 4 & 0.435 $\pm$ 0.011 \\
GatedGCN-E~\cite{bresson2017residual} & 4 & 0.282 $\pm$ 0.015 \\
RingGNN-E~\cite{chen2019equivalence} & 2 & 0.353 $\pm$ 0.019 \\
3WLGNN~\cite{maron2019provably} & 3 & 0.407 $\pm$ 0.028 \\
3WLGNN-E~\cite{maron2019provably} & 3 & 0.256 $\pm$ 0.054 \\  
GNNML3~\cite{balcilar2021breaking} & NA & 0.161 $\pm$ 0.006 \\
Graphormer~\cite{ying2021transformers} & NA & 0.122 $\pm$ 0.006 \\
CIN~\cite{bodnar2021weisfeiler2} & NA & \textbf{0.079} $\pm$ 0.006 \\ 	
ESAN~\cite{bevilacqua2022equivariant} & NA & 0.102 $\pm$ 0.003 \\
KP-GIN~\cite{feng2022powerful} & NA & 0.093 $\pm$ 0.007 \\
AgentNet~\cite{martinkus2023agent} & NA & 0.258 $\pm$ 0.033 \\
\midrule
PathNN-$\mathcal{SP}$ & 4 & 0.104 $\pm$ 0.002 \\
PathNN-$\mathcal{SP}^+$ & 4 & 0.131 $\pm$ 0.008 \\
PathNN-$\mathcal{AP}$ & 4 & 0.090 $\pm$ 0.004 \\
\bottomrule
\end{tabular}
\end{table}

Finally, we evaluate the proposed model in a graph regression task on ZINC$12$K in Table~\ref{tab:results_zinc}.
The three PathNN variants outperform most of the baselines on this dataset, despite, some of the baseline models, such as 3WLGNN, ESAN and GNNML3, being very expressive models considered to be state-of-the-art for many graph learning problems. 
With regards to the three PathNN variants, PathNN-$\mathcal{AP}$ performs best, but all three models exhibit promising performance.

\section{Conclusion}
We presented the PathNN model that aggregates path representations to generate node representations.
We proposed three different variants that focus on single shortest paths, all shortest paths and all simple paths of length up to $K$.
We have shown some of our PathNNs to be strictly more powerful than the $1$-WL algorithm.
Experimental results confirm our theoretical results.
The different PathNN variants were also evaluated on graph classification and graph regression tasks.
In most cases, our PathNNs outperform the baselines.

\section*{Acknowledgements}
G.N. is supported by ANR via the AML-HELAS (ANR-19-CHIA-0020) project and by the funding \'Ecole Universitaire de Recherche (EUR) Bertip, plan France 2030.

\bibliography{example_paper}
\bibliographystyle{icml2023}

\newpage
\appendix
\onecolumn

\section{Proofs of the Statements in Section \ref{sec:theory}}
\label{appendix:proofs}

\subsection{Proof of Theorem~\ref{theorem:simple_paths_wl}}

\begin{figure}[b]
    \centering
    \includegraphics[width=0.75\linewidth]{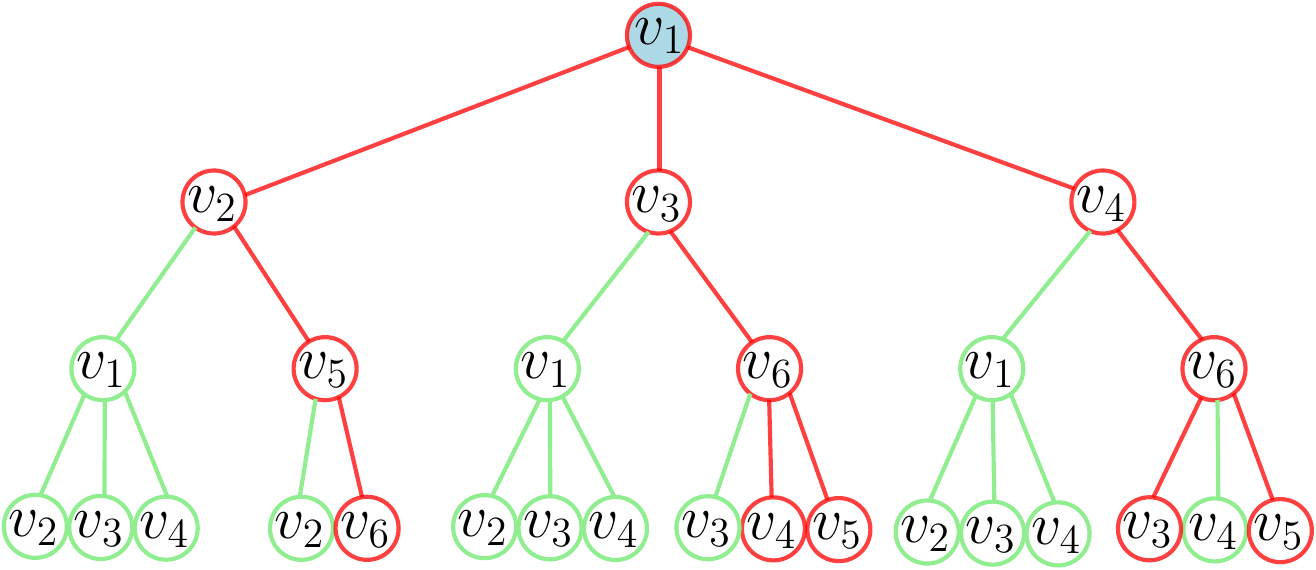}
    \caption{The WL tree rooted at $v_1$ of the graph presented in Figure~\ref{fig:graph}, $W_{v_1}^3$. \textcolor{red}{Incremental} nodes are marked in red, while \textcolor{green}{redundant} nodes are marked in green.} 
    \label{fig:subtree_decomposition}
\end{figure}
Let $W_v^k$ be the WL tree rooted at $v$ of height $k$ and $AP_v^k$ be the $\mathcal{AP}$-Tree rooted at $v$ of height $k$. We have that $W_v^k$ is an enumeration of walks of length $k$ that starts at $v$, and that $AP_v^k$ is an enumeration of simple paths of length $k$ that starts at $v$.

We start this proof by analyzing how nodes are distributed among levels of WL-Trees.
Nodes in a WL-Tree usually appear at multiple levels of the tree. 
If a node is part of level $l$ of the tree, it will always be part of levels $\{l+i : i \;\; \text{is\;even}\}$.
Besides, the same node can occur multiple times at the same level. 
On the other hand, Lemma \ref{lemma1} states that $AP_v^k$ is a subgraph of $W_v^k$. 
WL Trees thus contain elements that are not part of $\mathcal{AP}$-Trees. 
We define such elements as `redundant' where the terms comes from the fact that they have been processed on a higher level of the WL-Tree 
A `redundant' node $u$ at level $l$ of the tree is a node that either belongs to the set of its ascendants, or a node that has a `redundant' node in the set of its ascendants.
Ascendants of $u$ can be found by backtracking from $u$ up to the root node. 
Similarly, we define as `incremental' a node that is not `redundant'. 
With this definition, it is easy to reconstruct paths by only visiting `incremental' nodes when hopping from the root node to a leaf on $W_v^k$.
Similarly, walks can be reconstructed by visiting at least one `redundant' node from the top to the bottom of the tree. 

Since we can recover all paths using this definition, then we can see $AP_v^k$ as $W_v^k$ truncated from all its `redundant' nodes. Figure~\ref{fig:subtree_decomposition} provides an example based on the WL tree of the graph in Figure~\ref{fig:graph}, rooted at node $v_1$. 
Now that we have decomposed WL trees, let's see how the WL algorithm decides that two trees are structurally different. 

Suppose that at iteration $k-1$, the WL algorithm decides that $W_v^{k-1} = W_u^{k-1}$, but disambiguate the two trees at iteration $k$: $W_v^k \neq W_u^k$. 
Then there exists an `incremental' node, $v_i$, belonging to level $k-1$ of $W_v^{k-1}$ and an `incremental' node, $u_i$, belonging to level $k-1$ of $W_u^k$ such that $v_i$ and $u_i$ share similar higher level structure, but have different degree, $|\mathcal{N}(v_i)| \neq |\mathcal{N}(u_i)|$. 
The difference can only occur between ``incremental'' nodes $v_i$ and $u_i$ since `redundant' nodes have been processed on higher level of the tree $k' \in \{1,\dots, k-1\}$ and did not succeed to disambiguate $W_v^{k'}$ from  $W_u^{k'}$.
Besides, this structural difference must exist and must happen at level $k$ of the tree since $W_v^k \neq W_u^k$.  
In other words, it suffices to compare the `incremental' node distribution with respect to level $k$ of the trees of $\mathcal{N}(v_i)$ and $\mathcal{N}(u_i)$  to disambiguate $W_v^k$ from $W_u^k$. 
Comparing all `incremental' nodes included in $W_v^k$ to the ones included in $W_u^k$ is thus sufficient to distinguish WL trees, which can be done by comparing $AP_v^k$ to $AP_u^k$. 
Therefore, if $W_v^k \neq W_u^k$, then $AP_v^k \neq AP_u^k$. 

\begin{figure}
    \centering
    \includegraphics[width=.55\linewidth]{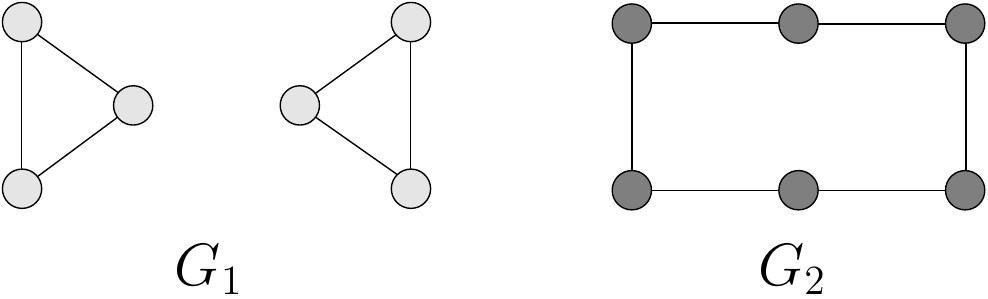}
    \caption{An example where PathNN can distinguish $G_1$ from $G_2$ using either $\mathcal{SP}, \mathcal{SP}^+, \mathcal{AP}$ but WL fails.}
    \label{fig:WL_fails}
\end{figure}

We now show that there exist pairs of graphs such that $AP_v^k \neq AP_u^k$ but $W_v^k = W_u^k$. We provide a simple example in Figure~\ref{fig:WL_fails} where taking simple paths of length $3$ is sufficient to distinguish $G_1$ from $G_2$, which concludes this proof. In particular, WL fails to distinguish $k$-regular graphs because of this tendency to capture redundant element, polluting the tree's structure. Instead, only capturing simple paths allows to remove all redundancy and thus to distinguish harder example of graphs. 

\subsection{Proof of Proposition~\ref{prop:SP_example}}

\begin{figure*}[t]
  \centering
  \subfigure[An open triplet (left) and a triangle (right).]{\includegraphics[width=.28\linewidth]{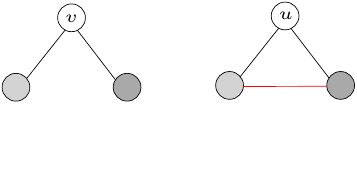}%
  \label{fig_first_case}}
  \hfill
  \subfigure[$W_v^2$ and $W_u^2.$]{\includegraphics[width=.28\linewidth]{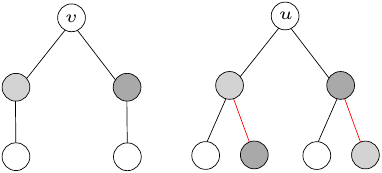}%
  \label{fig_second_case}}
  \hfill
  \subfigure[$P_v^2$ and $P_u^2$ for $P \in \{SP, SP^+\}.$]
  {\includegraphics[width=.28\linewidth]{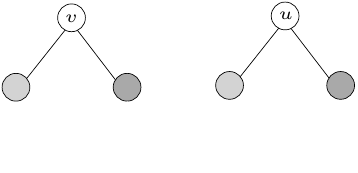}%
  \label{fig_third_case}}
  \caption{WL-Trees can distinguish node $v$ and $u$, while $\mathcal{SP}$-Trees and $\mathcal{SP}^+$-Trees fail to do so, since they do not include the `problematic edge' highlighted in \textcolor{red}{red}. However, $\mathcal{SP}$-Trees or $\mathcal{SP}^+$-Trees rooted at any node different than $u$ in the triangle graph will take the `problematic edge'  into account and hence guarantee disambiguation at the graph level. }
  \label{fig:example_proof_45}
\end{figure*}

To prove Proposition \ref{prop:SP_example} it suffices to show a counterexample for which $W_v^k$ and $W_u^k$ are structurally different and $P_v^k$ are identical $P_u^k$. Nodes $v$ and $u$ in Figure \ref{fig:example_proof_45}(a) have the required property, which concludes this proof. 

\subsection{Proof of Theorem~\ref{thm:ExpressivityOfAnnotatedSetsOfPaths}}
We will show that if $\tilde{P}_v^{k} = \tilde{P}_u^{k}$ holds for two nodes $v,u \in V$, then $W_v^{k} = W_u^{k}$ also holds.
We will prove this by induction on the depth $k$.
We begin with the case where $k = 1$.
Since $\tilde{P}_u^{1}$ contains as many paths as node $u$ has neighbours, $\tilde{P}_u^{1} = \tilde{P}_v^{1}$ implies that $W_u^{1} = W_v^{1}.$ 
Note that for all types of paths and for any length $k > 0$, the second node encountered when traversing the path is a neighbor of the source node, \ie $\pi(2) \in \mathcal{N}(v)$ for all $\pi \in \mathcal{P}_v$.
Then, it suffices to show that if $\Bigoms \left(\tilde{P}_{\pi(1)}^{k-1}, \tilde{P}_{\pi(2)}^{k-1}, \ldots, \tilde{P}_{\pi(k+1)}^{k-1}\right) \colon \pi \in \tilde{\mathcal{P}}_v^{k} \Bigcms = \Bigoms \left(\tilde{P}_{\pi(1)}^{k-1}, \tilde{P}_{\pi(2)}^{k-1}, \ldots, \tilde{P}_{\pi(k+1)}^{k-1}\right) \colon \pi \in \tilde{\mathcal{P}}_u^{k} \Bigcms$ holds for two nodes $v,u \in V$, then $\oms W_w^{k-1} \colon w \in \mathcal{N}(v) \cms = \oms W_w^{k-1} \colon w \in \mathcal{N}(u) \cms$.
For a contradiction, let us assume that $\oms W_w^{k-1} : w \in \mathcal{N}(v) \cms \neq \oms W_w^{k-1} \colon w \in \mathcal{N}(u) \cms$.
This can happen if the two multisets have different cardinalities or if $\oms W_w^{k-1} \colon w \in \mathcal{N}(v) \cms$ contains an element that is not contained in $\oms W_w^{k-1} : w \in \mathcal{N}(u) \cms$.
However, $\tilde{P}_v^{1} = \tilde{P}_u^{1}$, thus the two multisets have identical cardinalities.
Furthermore, the two multisets contain the same elements since $\oms \tilde{P}_{\pi(2)}^{k-1} \colon \pi \in \tilde{\mathcal{P}}_v^{k} \cms = \oms \tilde{P}_{\pi(2)}^{k-1} \colon \pi \in \tilde{\mathcal{P}}_u^{k} \cms$ (note that all the neighbors of the source node participate in at least one path since we extend terminating nodes with dummy nodes).
Thus, the only way in which the two multisets can be unequal is if the multiplicities of WL-Trees in $\oms W_w^{k-1} \colon w \in \mathcal{N}(v) \cms$  are different from their multiplicities in $\oms W_w^{k-1} \colon w \in \mathcal{N}(u) \cms$.
We will next show that this also cannot happen.
If the multiplicities of WL-Trees in the two multisets are different from each other, we can always find two nodes $w \in \mathcal{N}(v)$ and $z \in \mathcal{N}(u)$ such that $W_w^{k-1} \neq W_z^{k-1}$ while $W_w^{k-2} = W_z^{k-2}$.
We then consider nodes that occur as terminal nodes in paths where nodes $w$ and $z$ are involved, \ie $\oms \pi(k) \colon \pi \in \tilde{\mathcal{P}}_v^{k-1}, \pi(2)=w \cms$ and $\oms \pi(k) \colon \pi \in \tilde{\mathcal{P}}_u^{k-1}, \pi(2)=z \cms$.
These nodes have different number of children from each other since $W_w^{k-1} \neq W_z^{k-1}$.
Furthermore, for $k>2$, the annotations of nodes in the final positions of the annotated paths encode their degree.
Thus, we have that $\oms \pi(k) \colon \pi \in \tilde{\mathcal{P}}_v^{k-1}, \pi(2)=w \cms \neq \oms \pi(k) \colon \pi \in \tilde{\mathcal{P}}_u^{k-1}, \pi(2)=z \cms$ and therefore, $\tilde{P}_v^k \neq \tilde{P}_u^k$.
We have reached a contradiction and therefore, the multiplicities of WL-Trees of the two multisets are equal to each other.
Thus, we have that $\tilde{P}_v^{t} = \tilde{P}_u^{t} \implies W_v^{t} = W_u^{t}$.
Therefore, by contraposition, we have that $W_v^{t} \neq W_u^{t} \implies \tilde{P}_v^{t} \neq \tilde{P}_u^{t}$. 

To show the second component of our statement, that there are pairs of nodes $v, u$ for which $\tilde{P}_{v}^k \neq \tilde{P}_u^k$ while $W_{v}^k = W_{u}^k$, we simply need to consider one of the examples for which the WL test fails to distinguish two non-isomorphic graphs.
Figure \ref{fig:WL_fails} contains such a pair of graphs, which concludes our proof.

\section{Running Time}\label{sec:running_time}
We measured the running time of the proposed model on the DD (which contains the largest graphs among the considered TUDatasets) and ZINC12K datasets.
We use the same hyperparameter values for all models (\eg batch size, etc.).
We provide the running time of the proposed model and that of GIN (in seconds) in Table~\ref{tab:running_time}.
These experiments were run over an NVIDIA Tesla T4 GPU with 16GB of memory.
As expected, we observe that the running time of the proposed model is higher than that of GIN.
For $K=1$, the increase in the running time is not large, while no preprocessing takes place.
For $K > 1$, on DD, the extraction of the paths takes a significant amount of time (since the graphs contained in this dataset are large and dense), while on ZINC12K, extracting paths of length up to $4$ is relatively inexpensive (since graphs are small and sparse).
With regards to the increase in training and inference time, the most expensive model (PathNN-$\mathcal{AP}$) requires approximately $8$ times and $5$ times the time of GIN on DD and ZINC, respectively.
We conclude that, the experimental running times of the model for bounded values of $k$ are manageable.
On ZINC12K, we also measured the running time of a subgraph GNN model, DSS-GIN~\cite{bevilacqua2022equivariant}, and of KP-GIN~\cite{feng2022powerful} which are more expressive than GIN.
We can see that the preprocessing time of these two models is significantly greater than these of the proposed models.
Furthermore, training and inference time of KP-GIN is slightly smaller than those of the proposed models, while DSS-GIN is less efficient than PathNN-$\mathcal{SP}$ and PathNN-$\mathcal{SP}^+$, but slightly more efficient than PathNN-$\mathcal{AP}$.

\begin{table}[t]
\caption{Running time of the different models on the DD and ZINC12K datasets. All reported times are in seconds.}
\label{tab:running_time}
\centering
\scriptsize
\renewcommand{\arraystretch}{1.2}
\begin{tabular}{l|ccc}
\toprule
DD & Preprocessing & Time per Epoch (Training) & Inference Time \\
\midrule
GIN ($L=1$) & - & 0.36 ($\pm$ 0.03) & 0.03 ($\pm$ 0.00) \\
GIN ($L=2$) & - & 0.47 ($\pm$ 0.04) & 0.03 ($\pm$ 0.00) \\
\midrule
PathNN-$\mathcal{P}$ ($K=1$) & - & 0.56 ($\pm$ 0.02) & 0.05 ($\pm$ 0.00) \\
PathNN-$\mathcal{SP}$ ($K=2$) & {\color{white} 0}76.53 ($\pm$ 1.88) & 1.22 ($\pm$ 0.02) & 0.11 ($\pm$ 0.01) \\
PathNN-$\mathcal{SP}^+$ ($K=2$) & 139.17 ($\pm$ 2.38) & 1.57 ($\pm$ 0.02) & 0.13 ($\pm$ 0.01) \\
PathNN-$\mathcal{AP}$ ($K=2$) & {\color{white} 0}98.98 ($\pm$ 4.18) & 3.33 ($\pm$ 0.04) & 0.23 ($\pm$ 0.02) \\
\bottomrule
\end{tabular}\\
\vspace{.1cm}
\begin{tabular}{l|ccc}
\toprule
ZINC12K & Preprocessing & Time per Epoch (Training) & Inference Time \\
\midrule
GIN ($L=1$) & - & 0.90 ($\pm$ 0.08) & 0.13 ($\pm$ 0.03) \\
GIN ($L=4$) & - & 1.97 ($\pm$ 0.06) & 0.18 ($\pm$ 0.06) \\
DSS-GIN ($K=4$) & 112.27 ($\pm$ 4.17) & 4.25 ($\pm$ 0.11) & 0.25 ($\pm$ 0.05) \\
KP-GIN ($K=4$) & 197.47 ($\pm$ 0.57) & 3.09 ($\pm$ 0.05) & 0.13 ($\pm$ 0.03) \\
\midrule
PathNN-$\mathcal{P}$ ($K=1$) & - & 1.65 ($\pm$ 0.03) & 0.23 ($\pm$ 0.06) \\
PathNN-$\mathcal{SP}$ ($K=4$) & {\color{white} 0}21.28 ($\pm$ 0.36) & 3.56 ($\pm$ 0.04) & 0.26 ($\pm$ 0.06) \\
PathNN-$\mathcal{SP}^+$ ($K=4$) & {\color{white} 0}29.09 ($\pm$ 0.43) & 3.66 ($\pm$ 0.06) & 0.25 ($\pm$ 0.05) \\
PathNN-$\mathcal{AP}$ ($K=4$) & {\color{white} 0}25.93 ($\pm$ 0.89) & 4.38 ($\pm$ 0.06) & 0.32 ($\pm$ 0.05) \\
\bottomrule
\end{tabular}
\end{table}

As discussed above, the running time on DD is higher than that on ZINC12K since the graphs contained in DD are much larger than those contained in ZINC12K.
We provide in Table~\ref{tab:n_paths} the average number of paths of different type (\ie $\mathcal{SP}$, $\mathcal{SP}^+$ and $\mathcal{AP}$) and of different length for the two considered datasets.
We observe that the average number of paths for DD is much larger than that of the other dataset.
This justifies the difference in running time observed between the two datasets.

\begin{table}[t]
\caption{Average number of paths per graph on the DD and ZINC12K datasets. We do not report the number of paths of length higher than $2$ on DD since we did not measure the proposed model's running time for such lengths of paths.}
\label{tab:n_paths}
\centering
\scriptsize
\renewcommand{\arraystretch}{1.2}
\begin{tabular}{l|ccc|ccc}
\toprule
& \multicolumn{3}{c|}{DD} & \multicolumn{3}{c}{ZINC12K} \\
& $\mathcal{SP}$ & $\mathcal{SP}^+$ & $\mathcal{AP}$ & $\mathcal{SP}$ & $\mathcal{SP}^+$ & $\mathcal{AP}$ \\
\midrule
$K=2$ & 2,466 & 3,583 & 6,581 & 68.8 & 68.9 & 69.2 \\
$K=3$ & - & - & - & 68.1 & 79.0 & 88.0 \\
$K=4$ & - & - & - & 61.7 & 70.5 & 110.3 \\
\bottomrule
\end{tabular}
\end{table}

\section{Distance-Aware $\mathcal{AP}$-Trees}
\label{appendix:distance_enconding}
\begin{figure*}[h]
    \centering
    \subfigure[Two graphs.]{\includegraphics[width=0.45\linewidth]{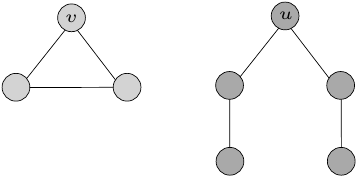}}
    \hfill
    \subfigure[$AP_v^2$ and $AP_u^2,$ respectively, augmented with distance encoding.]{\includegraphics[width=0.45\linewidth]{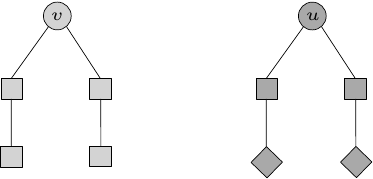}}
    \caption{Example of two nodes with similar $\mathcal{AP}$-Trees structures that can be distinguished with distance encoding. Node symbols in (b) corresponds to geodesic distance to the root node (squares for distance 1 and diamonds for distance 2).}
    \label{fig:example_de}
\end{figure*}

While $\mathcal{AP}$-Trees are more expressive than the 1-WL algorithm, they still fail in some cases where two nodes have similar $\mathcal{AP}$-Tree structures but play a different structural role. An example is provided in Figure~\ref{fig:example_de} where nodes $v$ and $u$ have similar $\mathcal{AP}$-Trees. However, at level $3$ of the tree rooted at $v$, only nodes with a geodesic distance of $1$ from $v$ are considered (Figure~\ref{fig:example_de}(b), left), while nodes with distance $2$ are considered at the same level of the tree rooted at $u$ (Figure~\ref{fig:example_de}(b), right). Using distance information in this situation allows us to disambiguate these two structurally similar trees, and therefore allows for more expressiveness.

We validate this theory empirically by noting that PathNN without the distance-aware LSTM cell can not distinguish any pair of graphs in SR datasets when considering simple paths of length up to $4$. On the other hand, PathNN with similar hyperparameters using the distance-aware LSTM cell is able to distinguish most pairs of graphs. 

\subsection{Distance Encoding}
\label{appendix:distance_encoding}
Shortest paths naturally encode the information of distance to the starting node.
However, simple paths lack this distance information as distance and position in simple paths are not directly related.
PathNNs directly take sequences of paths of increasing length as input and thus only have access to information over the sequence order. We found that including distance information when processing a node in the sequence is crucial for distinguishing harder examples of non-isomorphic graphs.
An example of how this process helps disambiguating Path-Trees is provided in Appendix~\ref{appendix:distance_enconding}.
We straightforwardly include distance information by designing a distance-aware LSTM cell.
We start by mapping distance to the start node to $d$-dimensional vectors using a learned embedding look-up table.
Let $\mathbf{d}_l \in \mathbb{R}^d$ be the representation of the distance value $l \in \{0,\dots,K\}$ and $d(v,u)$ be the geodesic distance between nodes $v$ and $u$. $\mathbf{d}_l$, $l \in {0,\dots,K}$ are vectors of weights, randomly initialized that will be learned during training to produce rich representation of distances to the central node.
Distance representations are shared across nodes, such that two pairs of nodes have equal distance representations if they have equal geodesic distance $l.$
We modify the usual LSTM cell to include distance information as follows
\begin{align*}
    l &= d \left( \pi(1), \pi(t) \right),\\
    \mathbf{i}_t &= \sigma \left( \mathbf{W}_{ii}\mathbf{h}_{\pi(t)} + \mathbf{W}_{ih} \mathbf{h}_{t-1} + \mathbf{W}_{id} \mathbf{d}_l \right),\\
    \mathbf{f}_t &= \sigma \left( \mathbf{W}_{fi}\mathbf{h}_{\pi(t)} + \mathbf{W}_{fh} \mathbf{h}_{t-1} + \mathbf{W}_{fd} \mathbf{d}_l\right),\\
    \mathbf{g}_t &= \tanh \left( \mathbf{W}_{gi}\mathbf{h}_{\pi(t)} + \mathbf{W}_{gh} \mathbf{h}_{t-1} + \mathbf{W}_{gd} \mathbf{d}_l \right),\\
    \mathbf{o}_t &= \sigma \left( \mathbf{W}_{oi}\mathbf{h}_{\pi(t)} + \mathbf{W}_{oh} \mathbf{h}_{t-1} + \mathbf{W}_{od} \mathbf{d}_l \right),\\
    \mathbf{c}_t &= \mathbf{f}_t \odot \mathbf{c}_{t-1} + \mathbf{i}_t \odot \mathbf{g}_t,\\
    \mathbf{h}_t &= \mathbf{o}_t \odot \tanh (\mathbf{c}_t),
\end{align*}
where $\mathbf{h}_{t}$ is the LSTM hidden state at time $t$, $\mathbf{c}_t$ is the cell state at time $t$, $\odot$ is the Hadamard product and $\mathbf{i}_t$, $\mathbf{f}_t$, $\mathbf{g}_t,$ $\mathbf{o}_t$ are the input, forget, cell and output gates respectively and $\sigma$ is the sigmoid function.
The distance-aware LSTM cell operates on sequences of tuples containing a node representation and the embedding of its distance to the starting node.
The distance embedding learns to produce meaningful representations of distances that help distinguishing graphs.
Since shortest paths directly include this distance information, the distance-aware LSTM cell is only used when PathNNs operate on $\mathcal{AP}.$

\section{Hyperparameter Configuration}
\label{appendix:hyperparameters}
Table~\ref{tab:param_config} displays the hyperparameter configuration for experiments conducted on ogbg-molhiv, Zinc12K, peptides-function and peptides-structural. For all of these experiments, we use 2-layer MLPs  with BN as the $\phi$ function in Equation~(\ref{update_function_2}). Initial node and edge features are projected to the desired dimensional space using ogb Atom and Bound Encoders. 

We found that using a mean readout function led to a performance boost on ogbg-molhiv and peptides-functional, as previously reported by traditional GNNs. We also experimented with a mean path aggregation function that replaces the sum over path representations in Equation~(\ref{update_function}) with averaging over path representations. We found that using a mean aggregation function combined with a sum readout function led to better model performances on peptides-structural. Note that we remove the BN layer of Equation~(\ref{update_function}) when using a mean path aggregation function.

We employ two kinds of early stopping strategy. The first method, \textit{patience}, stops training when the validation score has not improved for $r$ rounds, where $r$ is a hyperparameter. The second strategy, \textit{lr}, is used in addition to a learning rate scheduler that reduces the learning rate if the validation score has not improved for $r$ rounds. The training is stopped when the learning rate reaches a minimum threshold value. 

\begin{table}[t]
\centering
\caption{Hyperparameters configuration for our various experiments. Results are aggregated across 10 seeds for ogbg-molhiv and ZINC, and across 4 seeds for both peptides datasets.}
\label{tab:param_config}
\scriptsize
\renewcommand{\arraystretch}{1.2}
\begin{tabular}{l|l|cccc}
\toprule
& & ogbg-molhiv & Zinc & Peptides-functional & Peptides-structural \\
\midrule
\multirow{10}{*}{$\mathcal{SP}$}
& Epochs & 200 & 1000 & 200 & 200 \\ 
& Learning Rate & 0.0005 & 0.001 &  0.001 & 0.001 \\
& Early Stopping & \{patience, 50\} & \{lr, $1^{-5}$\} & \{patience, 50\} & \{patience, 50\}\\
& Scheduler & - & $\{20, 0.5\}$ &$\{20, 0.5\}$ & $\{20, 0.5\}$ \\
& K & 2 & 4 & 8 & 4 \\
& Hidden Size & 128 & 152 & 128 & 144 \\
& Batch Size & 128 & 128 & 128 & 128 \\
& Dropout & 0.5 & 0 & 0 & 0 \\
& Readout & mean  & sum & mean & sum \\
& Path Agg & sum & sum & sum & mean \\
& \# Parameters & 306177 & 497041 & 510090 & 469163\\
\midrule
\multirow{10}{*}{$\mathcal{SP}^+$}
& Epochs & 200 & 1000 & 200 & 200 \\
& Learning Rate & 0.0005 & 0.001 & 0.001 & 0.001\\
& Early Stopping & \{patience, 50\} & \{lr, $1^{-5}$\} & \{patience, 50\} & \{patience, 50\}\\
& Scheduler & - & $\{20, 0.5\}$ &$\{20, 0.5\}$ & $\{20, 0.5\}$ \\
& K & 2 & 4 & 8 & 4 \\
& Hidden Size & 128 & 152 & 128 & 144\\
& Batch Size & 128 & 128 & 128 & 128\\
& Dropout & 0.5 & 0 & 0 & 0 \\
& Readout & mean & sum & mean & sum\\
& Path Agg & sum & sum & sum & mean \\
& \# Parameters & 306177 & 497041 & 510090 & 469163\\
\midrule
\multirow{10}{*}{$\mathcal{AP}$}
& Epochs & 200 & 1000 & 200 & 200\\
& Learning Rate & 0.0005 & 0.001 & 0.001 & 0.001\\
& Early Stopping & \{patience, 50\} & \{lr, $1^{-5}$\} &  \{patience, 50\} & \{patience, 50\}\\
& Scheduler & - & $\{20, 0.5\}$ &$\{20, 0.5\}$ & $\{20, 0.5\}$ \\
& K & 2 & 4 & 7 & 4 \\
& Hidden Size & 128 & 140 & 122 & 136\\
& Batch Size & 128 &  128 & 128 & 128 \\
& Dropout & 0.5 & 0 & 0 & 0 \\
& Readout & mean & sum & mean &  sum \\
& Path Agg & sum & sum & sum & mean \\
& \# Parameters & 372097 & 501621 & 494720 & 494915\\
\bottomrule
\end{tabular}
\end{table}

\section{Including Edge Features}
\label{appendix:edge_encoding}
Molecular datasets used in our experiments contain additional edge features. We create a modified LSTM cell that computes path representations using node and edge features, as well as distance embeddings when processing simple paths. Let $\mathbf{e}_{u,v}$ be the edge representation of the edge $(u,v)$. Similarly to the distance-aware LSTM cell presented in Section \ref{appendix:distance_encoding}, we define the edge LSTM cell as follow: 

\begin{align*}
    \mathbf{i}_t &= \sigma \left( \mathbf{W}_{ii}\mathbf{h}_{\pi(t)} + \mathbf{W}_{ih} \mathbf{h}_{t-1} + \mathbf{W}_{ie} \mathbf{e}_{\pi(t-1), \pi(t)} \right),\\
    \mathbf{f}_t &= \sigma \left(\mathbf{W}_{fi}\mathbf{h}_{\pi(t)} + \mathbf{W}_{fh} \mathbf{h}_{t-1} + \mathbf{W}_{fe} \mathbf{e}_{\pi(t-1), \pi(t)} \right),\\
    \mathbf{g}_t &= \tanh \left(\mathbf{W}_{gi}\mathbf{h}_{\pi(t)} + \mathbf{W}_{gh} \mathbf{h}_{t-1} + \mathbf{W}_{ge} \mathbf{e}_{\pi(t-1), \pi(t)} \right),\\
    \mathbf{o}_t &= \sigma \left(\mathbf{W}_{oi}\mathbf{h}_{\pi(t)} + \mathbf{W}_{oh} h_{t-1} + \mathbf{W}_{oe} \mathbf{e}_{\pi(t-1), \pi(t)} \right),\\
    \mathbf{c}_t &= \mathbf{f}_t \odot \mathbf{c}_{t-1} + \mathbf{i}_t \odot \mathbf{g}_t,\\
    \mathbf{h}_t &= \mathbf{o}_t \odot \tanh (\mathbf{c}_t).
\end{align*}

For $t=1$, we set $\mathbf{e}_{\pi(t-1), \pi(t)}$ to a vector of zeros as the edge LSTM cell starts by processing a node representation, and no connection to other nodes exists yet. For $t \geq 2$, the feature vector of the edge that connects the previous and current node in the path is processed together with the current node's representation. Note that we also use a version of this LSTM cell that also processes distance embeddings (as explained in Appendix \ref{appendix:distance_encoding}) when processing simple paths. 

\section{Datasets Details}
\label{appendix:datasets}
The CSL dataset contains $4$-regular graphs with edges connected to form a cycle and containing skip-links between nodes.
There are $150$ graphs in total, each consisting of $41$ nodes.
The $150$ graphs are equally divided into $10$ isomorphism classes based on the skip-link length of the graph.
EXP is a synthetic dataset that contains pairs of graphs that are not distinguishable by the $1$-WL algorithm.
However, these graphs are $2$-WL distinguishable.
Each graph encodes a propositional formula, and given a pair $G, G'$ of non-isomorphic graphs, the two graphs have different SAT outcomes, \ie $G$ encodes a satisfiable formula, while $G'$ encodes an unsatisfiable formula.
We used the dataset for both graph isomorphism testing and as a binary classification task.
SR is a dataset that contains strongly regular graphs.
Those graphs are not distinguishable by the $3$-WL algorithm.
A set of strongly regular graphs with parameters $(p_1,p_2,p_3,p_4)$ means that each graph has $p_1$ nodes, the degree of each node is equal to $p_2$, nodes that are connected by an edge share $p_3$ common neighbors and nodes that are not connected by an edge share $p_4$ common neighbors.

The ENZYMES dataset contains $600$ protein tertiary structures represented as graphs obtained from the BRENDA enzyme database.
Each enzyme is a member of one of the Enzyme Commission top level enzyme classes (EC classes) and the task is to correctly assign the enzymes to their classes \cite{borgwardt2005protein}.
The NCI1 dataset contains more than four thousand chemical compounds screened for activity against non-small cell lung cancer and ovarian cancer cell lines \cite{wale2008comparison}.
PROTEINS contains proteins represented as graphs where vertices are secondary structure elements and there is an edge between two vertices if they are neighbors in the amino-acid sequence or in $3$D space.
The task is to classify proteins into enzymes and non-enzymes \cite{borgwardt2005protein}. 
D\&D is a dataset that contains over a thousand protein structures.
Each protein is a graph whose nodes correspond to amino acids and a pair of amino acids are connected by an edge if they are less than $6$ \AA ngstroms apart.
The task is to predict if a protein is an enzyme or not \cite{dobson2003distinguishing}.
The IMDB-B and IMDB-M datasets were created from IMDb, an online database of information related to movies and television programs. 
The graphs contained in the two datasets correspond to movie collaborations.
The vertices of each graph represent actors/actresses and two vertices are connected by an edge if the corresponding actors/actresses appear in the same movie.
Each graph is the ego-network of an actor/actress, and the task is to predict which genre an ego-network belongs to \cite{yanardag2015deep}.

The ogbg-molhiv dataset is a molecular property prediction dataset that is adopted from the MoleculeNet~\cite{wu2018moleculenet}.
The dataset consists of $41,127$ molecules and corresponds to a binary classification dataset where the task is to predict whether a molecule inhibits HIV virus replication or not.
The molecules in the training, validation and test sets are divided using a scaffold splitting procedure that splits the molecules based on their two-dimensional structural frameworks.
The scaffold splitting attempts to separate structurally different molecules into different subsets.

ZINC is one of the most popular molecular datasets where the task is to predict the constrained solubility of molecules, an important chemical property for designing generative GNNs for molecules.

Peptides-func and Peptides-struct datasets are derived from $15,535$ peptides retrieved from SATPdb database that includes the
sequence, molecular graph, function, and 3D structure of the
peptides.
Both datasets use the same set of graphs but differ in their prediction tasks.
Peptides-func is a multi-label graph classification dataset where there exist $10$ classes in total associated with the peptide function.
Peptides-struct is a multi-label graph regression dataset where the goal is to predict aggregated 3D properties of the peptides at the graph level.

\end{document}